\documentclass[letterpaper, 10 pt, conference]{ieeeconf}  

\IEEEoverridecommandlockouts                              

\usepackage{graphicx} 
\usepackage{subcaption}
\usepackage{amsmath} 
\usepackage{bm} 
\usepackage{url}
\usepackage{hyperref}
\usepackage[usenames,dvipsnames,table]{xcolor}

\usepackage{booktabs}
\usepackage{xspace}
\usepackage{multirow}
\usepackage{amsfonts}
\usepackage{makecell}
\usepackage[table]{xcolor}
\definecolor{lightgray}{rgb}{0.9,0.9,0.9}
\newcommand\blfootnote[1]{%
\begingroup
\renewcommand\thefootnote{}\footnote{#1}%
  \addtocounter{footnote}{-1}%
  \endgroup
}

\makeatletter
    \let\NAT@parse\undefined
\makeatother
\usepackage[square,numbers,sort&compress]{natbib}

\usepackage{hyperref}
\hypersetup{
    colorlinks=true,
    linkcolor=MidnightBlue,
    filecolor=magenta,      
    urlcolor=MidnightBlue,
    citecolor=MidnightBlue,
}
\usepackage[capitalise, nameinlink]{cleveref}

\begin{document}
\bstctlcite{IEEEexample:BSTcontrol}

\title{\LARGE \bf MOVE: A Simple Motion-Based Data Collection Paradigm for Spatial Generalization in Robotic Manipulation}

\author{Huanqian Wang$^{*,1}$, Chi Bene Chen$^{*,1}$, Yang Yue$^{*,\S,1}$, Danhua Tao$^3$, Tong Guo$^1$, Shaoxuan Xie$^2$\\ Denghang Huang$^2$, Shiji Song$^1$, Guocai Yao$^{\dagger\,2}$, Gao Huang$^{\dagger\,1}$}

\maketitle

\begin{abstract}

Imitation learning method has shown immense promise for robotic manipulation, yet its practical deployment is fundamentally constrained by the data scarcity. Despite prior work on collecting large-scale datasets, there still remains a significant gap to robust spatial generalization. 
We identify a key limitation: individual trajectories, regardless of their length, are typically collected from a \emph{single, static spatial configuration} of the environment. This includes fixed object and target spatial positions as well as unchanging camera viewpoints, which significantly restricts the diversity of spatial information available for learning.
To address this critical bottleneck in data efficiency, we propose \textbf{MOtion-Based Variability Enhancement} (\emph{MOVE}), a simple yet effective data collection paradigm that enables the acquisition of richer spatial information from dynamic demonstrations.
Our core contribution is an augmentation strategy that injects motion into any movable objects within the environment for each demonstration. This process implicitly generates a dense and diverse set of spatial configurations within a single trajectory. 
We conduct extensive experiments in both simulation and real-world environments to validate our approach. For example, in simulation tasks requiring strong spatial generalization, \emph{MOVE} achieves an average success rate of 39.1\%, a 76.1\% relative improvement over the static data collection paradigm (22.2\%), and yields up to 2--5$\times$ gains in data efficiency on certain tasks.
Our code is available at \url{https://github.com/lucywang720/MOVE}$^4$.

\end{abstract}

\blfootnote{\hspace{-2mm} $^*$Equal contribution. $\S$Project lead. $^\dagger$Corresponding author. }
\blfootnote{\hspace{-2mm} $^1$BNRist, Tsinghua University. $^2$Beijing Academy of Artificial Intelligence. $^3$Southeast University. }
\blfootnote{\hspace{-2mm} $^4$The real-world dataset is available at \url{https://huggingface.co/datasets/BAAI/MOVE}.}


\section{Introduction}

Recently, end-to-end learning methods have made significant strides in robotic control, enabling the completion of numerous complex manipulation tasks. 
The state-of-the-art approaches, exemplified by Diffusion Policy~\cite{chi2024diffusionpolicy, liu2025rdt1bdiffusionfoundationmodel} and Vision-Language-Action models~\cite{octomodelteam2024octoopensourcegeneralistrobot, cheang2024gr2generativevideolanguageactionmodel, black2024pi0visionlanguageactionflowmodel}, leverage large-scale datasets to achieve impressive generalization capabilities across different objects, new tasks, and varying environments~\cite{10611331, teoh2024green}. 
These advancements mark a significant step towards general-purpose embodied intelligence. Despite these achievements, generalization across spatial variations in object pose remains a critical yet overlooked challenge~\cite{xing2021kitchenshift, tsagkas2025pretrainedvisualrepresentationsfall}. This limitation is particularly acute for real-world deployment, where robots must operate in unstructured environments far more variable than the controlled settings in simulation environments.

\begin{figure}[h]
\centering
\begin{minipage}{0.235\textwidth}
\centering
\includegraphics[width=1\textwidth]{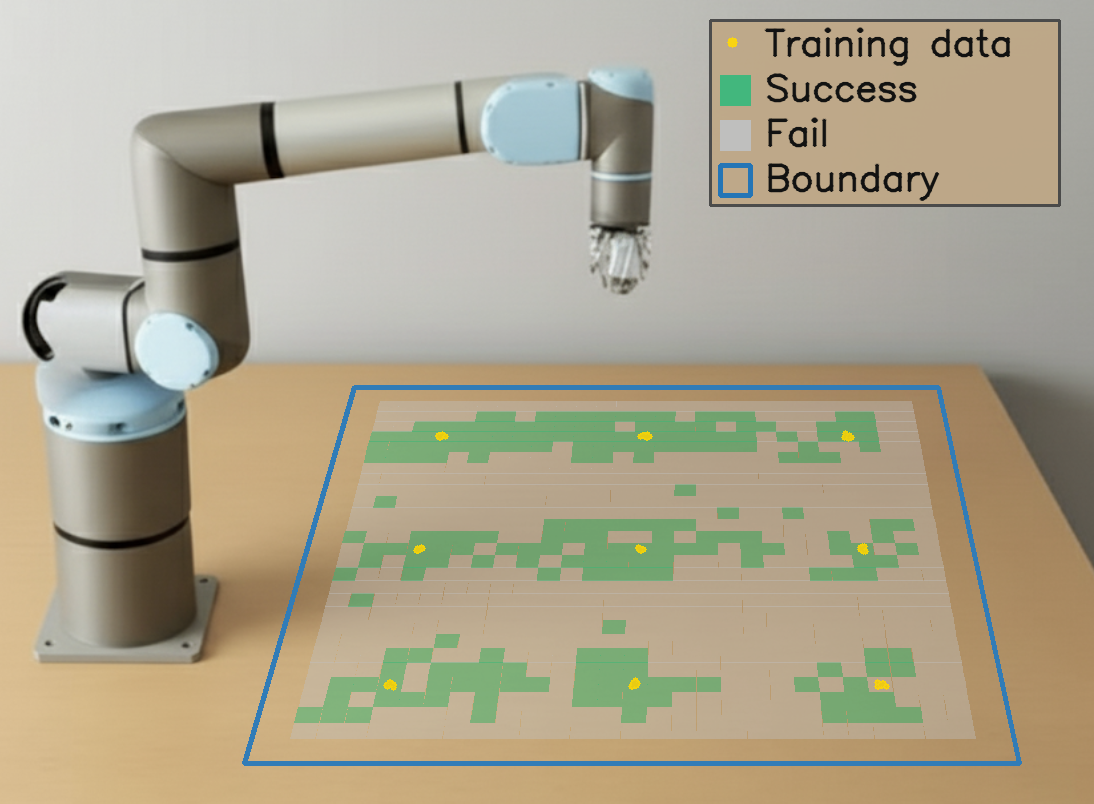}
\end{minipage}\hfill
\begin{minipage}{0.235\textwidth}
\centering
\includegraphics[width=1\textwidth]{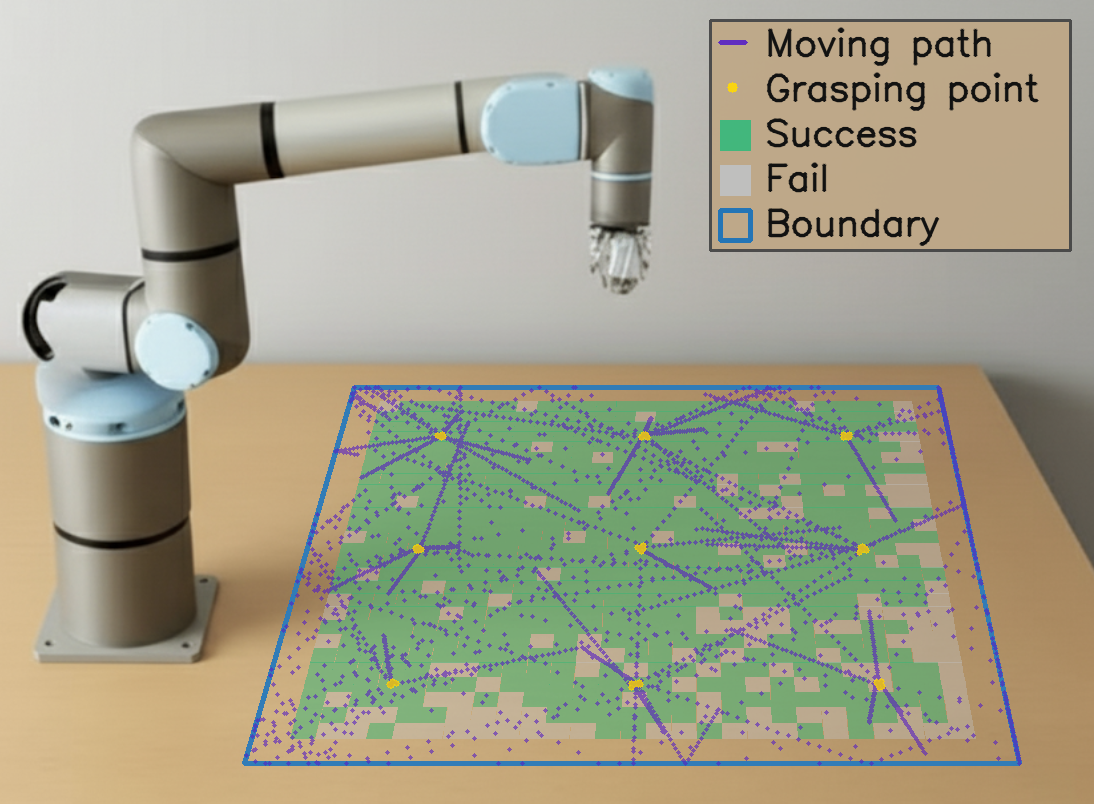}
\end{minipage}\hfill 
\caption{
We uniformly sample 10 trajectories from each of the 9 points across the entire space using both static data collection and MOVE.
To ensure a fair comparison, we enforce that the grasping point of each MOVE trajectory corresponds to that of a static trajectory and the same total number of timesteps.
Despite this alignment, MOVE exhibits significantly better spatial generalization to unseen grasp points (29.5\% vs. 80.8\%).}
\label{fig:pre1}
\end{figure}

The root of this problem is the inefficient sampling of spatial configuration from a continuous state space by static data collection methods.
We highlight the limitations of static data collection in Figure~\ref{fig:pre1} left. We trained a diffusion policy using data uniformly gathered from 9 spatial locations and evaluated the policy across the entire object space. The policy, as expected, succeeds only around the locations in the training set, but fails at other test points in much of the remaining space, resulting in a success rate of 29.5\%.
This issue becomes especially severe as the spatial dimensionality of the task space increases. 
For instance, diverse camera perspectives, adjustable table heights, and randomized target object placements contribute to a more complex and combinatorially rich spatial setting.
As shown in Table~\ref{table:intro}, success rates decay exponentially as the spatial dimensions expand, revealing the poor generalization capabilities of current methods in real-world environments.

To tackle this problem, we challenge the paradigm of static data collection itself. 
In this paradigm, an entire expert trajectory, often spanning several hundred timesteps, captures a task under a fixed spatial configuration, such as a fixed object position, target position, and camera viewpoint.
Spatial sparsity leads to the consequence that if the policy needs to grasp an object in a new pose, completely new demonstrations must be collected for that specific location~\cite{xue2025demogensyntheticdemonstrationgeneration, tan2024maniboxenhancingspatialgrasping}, which is impractical for real-world deployment.


\begin{figure*}[h]
    \centering
    \begin{subfigure}{0.95\textwidth} 
        \centering
        \includegraphics[width=\linewidth]{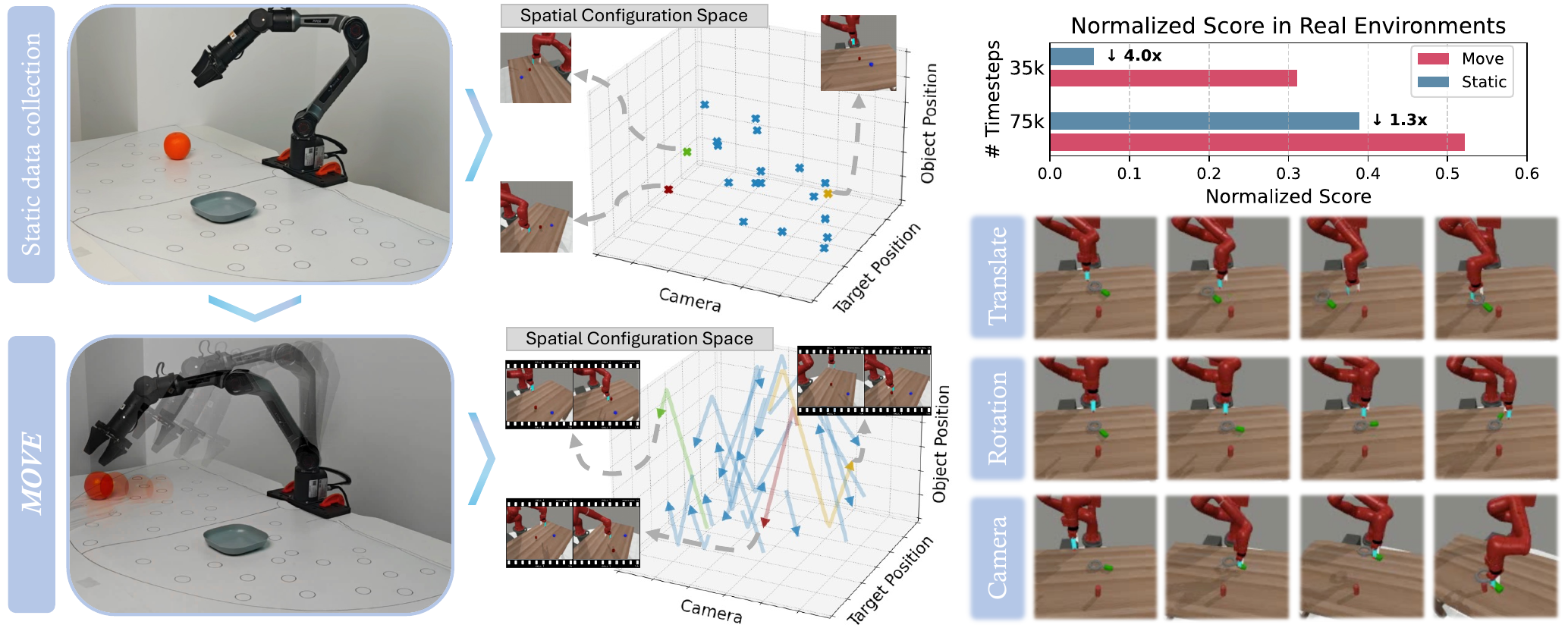}
    \end{subfigure}
    \caption{
    \textbf{An overview of the MOVE data collection paradigm.}
    (\textbf{Left}) 
    A conceptual comparison between the standard static data collection paradigm and \emph{MOVE}.  
    The former samples from discrete, fixed spatial configurations, where each trajectory represents a single point in the spatial configuration space.  
    In contrast, each trajectory collected by \emph{MOVE} is treated as a continuous segment, with objects, targets, and the camera in motion, resulting in a dense and diverse set of spatial configurations within a single trajectory.
    Therefore, with the same number of trajectories, our approach encodes broader spatial coverage and richer spatial information. 
    (\textbf{Right Top}) In real-world environments, policies trained with data collected via \emph{MOVE} demonstrate superior performance and generalization compared to the traditional data collection paradigm, with a maximum improvement of 4.0x on the normalized score. 
    (\textbf{Right Bottom}) We demonstrate several forms of motion augmentation employed in \emph{MOVE}, including translation, rotation and camera motion. }
    \label{fig:demo}
    
\end{figure*}



In this work, we introduce \emph{MOVE}, a motion-based data collection framework which enhances the spatial information density per trajectory to improve the spatial generalization in robotic manipulation. Motivated by limitations of traditional data collection, we aim to endow a single trajectory with spatial location information from more than just one spatial configuration. Specifically, the key objects, such as pickup object, target object and camera, are intentionally and continuously moved when collecting expert human demonstrations.
We illustrate the spatial coverage of \emph{MOVE} compared to static methods in Figure~\ref{fig:demo}.
Given an equivalent dataset volume, \emph{MOVE} exposes the policy to a stream of continuously moving objects within a single trajectory during training. 
Although the policy may not directly learn to grasp objects at every position along the motion path, it can still implicitly acquire knowledge of the corresponding spatial configurations. 
As shown in Figure~\ref{fig:pre1} (right), a policy trained with \emph{MOVE} data is able to grasp the object along its motion path.
This paradigm effectively embeds a powerful and flexible form of data augmentation directly into each trajectory, thereby enhancing both sample efficiency and 
spatial generalization.






We validate the effectiveness of \emph{MOVE} through extensive experiments in both simulation and real-world scenarios and demonstrate that a policy trained with \emph{MOVE} data collection paradigm evidently improves data efficiency and achieves better spatial generalization.
In the Meta-World simulation environments requiring strong spatial generalization, \emph{MOVE} on average achieves a success rate of 39.1\%, a 76.1\% improvement over the static data collection paradigm (22.2\%).
For example, in the Pick-and-Place task, a 20k-timestep \emph{MOVE} dataset matches the success rate of a 50k-timestep static dataset, and in the Assembly task, a 50k-timestep \emph{MOVE} dataset achieves comparable performance to a 100k static dataset. 
Furthermore, in real-world experiments with highly randomized spatial configurations that pose a substantial generalization challenge, \emph{MOVE} achieves a 23.3\% success rate with only 35k timesteps, dramatically outperforming the static method under the same data budget (3.3\% at 35k) and matching its performance of the static method that is trained with more than twice the data (23.3\% at 75k).
These results highlight that \emph{MOVE} substantially reduces the amount of data required to achieve the same level of spatial generalization, demonstrating dynamic data collection paradigm's potential to enable spatial generalization more effectively and efficiently that supports scalable learning.


\section{Related Works}


\subsection{Robotic manipulation and spatial generalization}
Robot manipulation has recently made significant progress, where policies represented by Vision-Language-Action (VLA) models and diffusion models enable robots to perform a wide range of tasks based on visual inputs. Built upon Vision-Language Models, VLA models utilize large pretrained transformers to map visual and linguistic inputs to robotic actions.
~\cite{brohan2023rt2visionlanguageactionmodelstransfer, kim2024openvlaopensourcevisionlanguageactionmodel, cheang2024gr2generativevideolanguageactionmodel, octomodelteam2024octoopensourcegeneralistrobot, black2024pi0visionlanguageactionflowmodel,yue2024deer}. Diffusion Policy~\cite{chi2024diffusionpolicy} and other extensions~\cite{ze20243ddiffusionpolicygeneralizable, liu2025rdt1bdiffusionfoundationmodel, ke20243d, wen2025dexvla, ma2024hierarchical}, leverage the capabilities of diffusion models to fit multi-modal action distributions and enhance long-horizon planning and efficiency by using action chunking, demonstrated remarkable success in learning complex, dexterous skills directly from human demonstrations.

Despite these advancements, achieving spatial generalization is still a central and long-term challenge in robotics. Many prior works attempted to solve this problem by improving visual representations through injecting spatial information such as 3D point clouds or bounding box~\cite{zhu2024spa, qu2025spatialvlaexploringspatialrepresentations, li2024generalistrobotpoliciesmatters, tan2024maniboxenhancingspatialgrasping}. 
Nevertheless, the performance of robotic policies is still critically dependent on the scale and diversity of their training datasets~\cite{10611331, lin2025datascalinglawsimitation}, and degrades severely especially when the test scenarios are different from the training distribution. As a result of this dependency, the field is shifting from a purely model-centric to a data-centric viewpoint~\cite{bu2025agibot, khazatsky2025droidlargescaleinthewildrobot}. 
To this end, researchers are dedicated to addressing the challenge that collecting large-scale, real-world robotic data is resource-intensive and time-consuming.

\subsection{Data Collection in Robotics}


Inspired by the success of data scaling in LLMs and VLMs, researchers have also begun exploring data scaling for manipulation tasks.	
The development of large-scale datasets has been instrumental to recent progress, including DROID (76k trajectories)~\cite{khazatsky2025droidlargescaleinthewildrobot}, BridgeData V2 ($\sim$60k trajectories)~\cite{walke2024bridgedatav2datasetrobot} and the Open X-Embodiment dataset ($\sim$1M trajectories)~\cite{embodimentcollaboration2025openxembodimentroboticlearning}.
Despite significant efforts from the community, the quantity of available robot data remains far below that of vision-language data, limiting the ability of current methods to achieve robust generalization~\cite{zhong2025surveyvisionlanguageactionmodelsaction}.
To address this challenge, recent research investigated more efficient data acquisition methods, following directions including large-scale physical simulation and 3D scene reconstruction.
Simulation-based methods collect extensive data in high-fidelity simulations to bridge the sim-to-real gap~\cite{zhu20243dgaussiansplattingrobotics, mu2025robotwindualarmrobotbenchmark, mu2021maniskillgeneralizablemanipulationskill, do2025watchlessfeelmore}.
3D reconstruction-based methods can generate synthetic trajectories based on real trajectories~\cite{mandlekar2023mimicgendatagenerationscalable, xue2025demogensyntheticdemonstrationgeneration}.
Beyond these methods, researchers are also exploring efficient strategies for collecting real-world data. ADC~\cite{huang2025adversarialdatacollectionhumancollaborative} shares some similarities with our work, as it periodically resets the object's position during data collection. However, ADC includes only a few discrete points along a trajectory, whereas \emph{MOVE} captures richer spatial information by forming a continuous curve in the location space. Moreover, \emph{MOVE} naturally extends to additional spatial dimensions, allowing the incorporation of variables such as camera motion and dynamic table height.

\section{Approach}

\subsection{Challenge in Spatial Generalization}

In robot learning, the generalization of policy training heavily relies on large, diverse datasets, but acquiring enough data in robotics is notably difficult to achieve. 
This challenge becomes more pronounced as the spatial dimensionality of the task increases, since each standard demonstration typically contains only a single instance of a spatial configuration, leading to severe spatial sparsity in high-dimensional environments.
We validate this phenomenon on the Metaworld Pick-Place task and list the results in Table~\ref{table:intro}.

Specifically, we construct three settings of increasing difficulty for spatial generalization by progressively randomizing key spatial factors.
In setting 1, Only the object's position is randomly initialized within a $30\,\text{cm} \times 30\,\text{cm}$ area on the table, while the target position and camera viewpoint remain fixed. Although training and testing use the same sampling range, generalization is required due to the difference in sampled positions.
In setting 2, in addition to randomizing the object’s position, the target position is also randomized within a $20\,\text{cm} \times 10\,\text{cm} \times 25\,\text{cm}$ volume.
Setting 3 further increases difficulty by randomizing the camera pose, with the viewing angle ranging from $0$ to $\pi$ radians, transitioning from a controlled setup to a more realistic scenario.
Under the same training budget of 20k timesteps, the success rate drops dramatically from 67.5\% in Setting 1 to 31.7\% in Setting 3. This exponential decline highlights the inability of the standard static data collection paradigm to sufficiently cover the spatial configuration space in realistic environments with multi-dimensional variation.

\textbf{Overview.}\quad To enhance the model's generalization to complex environments and improve the robustness against unforeseen spatial positions, we explore a simple yet effective data collection approach termed \textbf{MOtion-Based Variability Enhancement} (\emph{MOVE}), which leverages dynamic trajectories to provide richer spatial grounding signals. 
As Figure~\ref{fig:demo} shows, to increase the coverage of spatial configurations, we introduce controlled kinematic motions, including translation, rotation, and camera movement, as a data augmentation strategy when collecting training data (Section~\ref{method:dyna}). Once collected, we apply diffusion policy to train policy models (Section~\ref{method:dp}).

\begin{table}[!htbp]
\centering
\caption{
\textbf{Impact of Real-World Spatial Variation on Success Rate.} 
We progressively introduce variations in object placement, target placement, and camera viewpoint. 
Unlike simulation settings where these factors are typically fixed or only slightly perturbed, each training and testing trajectory is collected under \emph{a randomized configuration} sampled from the same distribution.}
\label{table:intro}
\begin{tabular}{cccc}
\toprule
\multirowcell{2}{Randomized\\Factors}
  & \multicolumn{3}{c}{From research setting $\Longrightarrow$ real world} \\
\cmidrule(lr){2-4}
    & Object & + Target & + Camera \\
\midrule
Success rate & 0.675 & 0.447 & 0.317 \\
\bottomrule
\end{tabular}
\end{table}

\subsection{Spatial Configuration Augmentation}
\label{method:dyna}



\textbf{Object Translation:} 
To ensure spatial coverage of the entire workspace, \emph{MOVE} simulates multiple linear motion trajectories for \emph{both the pickup and target objects}, which is modelled by incorporating linear translation and bounce at the boundaries. An object's position $\mathbf{p}_i(t)$ at time $t$ is determined by its initial position $\mathbf{p}_i(0)$, a constant velocity vector $v_i$, and moving direction $\mathbf{d}_i$.
 


\begin{equation}
\begin{aligned}
\mathbf{p}_i(t) &= \mathbf{p}_i(0) + t \cdot v_i \cdot v_{\max} \cdot \mathbf{d}_i, \quad \forall i \in \{\text{pick}, \text{target}\} \\ 
&\text{with} \quad v_i \sim \mathrm{B}(\alpha_p, \beta_p), \quad \mathbf{d}_i \sim U(\mathbb{S}^2) 
\end{aligned}
\end{equation}

where $v_{\max}$ is the maximum possible speed and velocity $v_i$ is sampled from a Beta distribution $\mathrm{B}(\alpha_p, \beta_p)$. We leverage Beta distribution's properties to constrain the sampled velocity to the interval $\text{[}0,v_{\max} \text{]}$ and ensure a higher probability of speeds approaching 0 than $v_{\max}$. Specifically, we set all $\alpha=2$ and $\beta=5$ throughout our simulation experiments. This distribution facilitates the model's learning of not only spatial generalization but also robust grasping strategies.

\textbf{Object Rotation:} To improve spatial generalization to a wide range of object orientations, we introduce constant angular velocity rotations. For simplicity, we model 1-D rotation around the vertical z-axis. Similarly to the above, the orientation $\theta_i(t)$ of an object evolves based on its initial orientation $\theta_i(0)$ and a constant angular velocity $\omega_i$.


\begin{equation}
\begin{aligned}
\bm{\theta}_i(t) &= \bm{\theta}_i(0) + t \cdot \omega_i \cdot \omega_{\max} \cdot \mathbf{d}_i, \quad \forall i \in \{\text{pick}, \text{target}\} \\ 
&\text{with} \quad \omega_i \sim \mathrm{B}(\alpha_{\theta}, \beta_{\theta}), \quad \mathbf{d}_i \in \{-1, 1\}  
\end{aligned}
\end{equation}

This augmentation is particularly beneficial for asymmetric objects such as mugs with handles. In contrast, it is not applied to objects that are fully rotationally symmetric.

\textbf{Camera Movement:} 
To simulate a non-static viewpoint, the virtual camera moves along a constrained cylindrical path relative to the scene's center.
The camera's position is updated similarly to object translation, with its velocity sampled from a Beta distribution.

\begin{equation}
\begin{aligned}
\mathbf{p}_i(t) &= \mathbf{p}_i(0) + t \cdot u_i \cdot u_{\max} \cdot \mathbf{d}_i, \quad \forall i \in \{\text{camera}\} \\ 
&\text{with} \quad u_i \sim \mathrm{B}(\alpha_c, \beta_c), \quad \mathbf{d}_i \sim U(\mathbb{S}^2)  
\end{aligned}
\end{equation}

\textbf{Combined Augmentation Strategy:} 
Rather than applying all motions simultaneously, we employ a staged strategy tailored to the semantic phases of a task. 
For example, in the Box-Close task, we decompose each trajectory into the pick phase ($t_0\rightarrow t_1$) and a placement phase ($t_1\rightarrow t_2$).
\begin{itemize}
    \item Pick phase ($t_0\rightarrow t_1$): we apply translation and rotation only to the pickup object (the box lid), the camera movement is also introduced. This forces the policy to learn to approach and grasp a moving target while adapting to a changing viewpoint.
    \item Placement phase ($t_1\rightarrow t_2$): we apply linear translation only to the target object (the box body) and continue the dynamic camera motion. This challenges the policy to place the object onto a moving destination.
\end{itemize}

This strategy expands \emph{MOVE} beyond a single dimension, increasing the spatial information richness across multiple spatial dimensions. We validate this combined augmentation strategy in Section~\ref{sec:abla}.

\subsection{Training}
\label{method:dp}

To learn the robot's control strategy, we adopt the Diffusion Policy~\cite{chi2024diffusionpolicy} framework. 
For training, we utilize the dynamic dataset collected following the methodology described in Section~\ref{method:dyna}.
Specifically, we employ the Denoising Diffusion Implicit Models (DDIM) scheduler for a deterministic and efficient sampling process.The less noisy sample $x_{t-1}$ is computed from the noised sample $x_t$ at timestep $t$ as follows:
\begin{equation}
\begin{aligned}
        \mathbf{x}_{t-1} &= \sqrt{\bar{\alpha}_{t-1}} \left( \frac{\mathbf{x}_t - \sqrt{1-\bar{\alpha}_t} \boldsymbol{\epsilon}_\theta(\mathbf{x}_t, t)}{\sqrt{\bar{\alpha}_t}} \right) \\
        &+ \sqrt{1-\bar{\alpha}_{t-1}} \cdot \boldsymbol{\epsilon}_\theta(\mathbf{x}_t, t)
\end{aligned}
\end{equation}
A comprehensive list of hyperparameters, architecture details, and other settings can be referred to in the Appendix~\ref{appendix}.

\section{Experiments}

\subsection{Preliminary Experiment}

We begin with preliminary experiments designed to showcase the effectiveness of \emph{MOVE} in achieving better spatial generalization under a clean and controlled setting in the Pick-Place simulation task.  
Specifically, we define a set of discrete grasping points and ensure that both the static data collection method and \emph{MOVE} perform grasps at these same locations during data collection.  
In the \emph{MOVE} setting, objects are initialized at random positions and then move toward the predefined grasping points.  
We also keep the total number of timesteps for training the same across both methods.

\paragraph{Generalization Comparison from Sparse Sampling} We uniformly choose 9 points across the space. For each point, we sample 10 trajectories for static data collection as we empirically find that sparse sampling with less trajectories usually failed to learn a meaningful policy. We sampled total 85 trajectories for \emph{MOVE} to ensure the same total number of timesteps.
We evaluate both policies across the entire space and plot the results in Figure~\ref{fig:pre1}. Although the policy trained on static data performs well on the training points, \emph{MOVE} demonstrates superior generalization to the entire space, succeeding even in locations far from training points. Specifically, \emph{MOVE} achieves a global success rate of 80.8\% across the entire space, significantly outperforming the static policy's 29.5\%.

\begin{figure}[h]
\centering
\begin{minipage}{0.235\textwidth}
\centering
\includegraphics[width=1\textwidth]{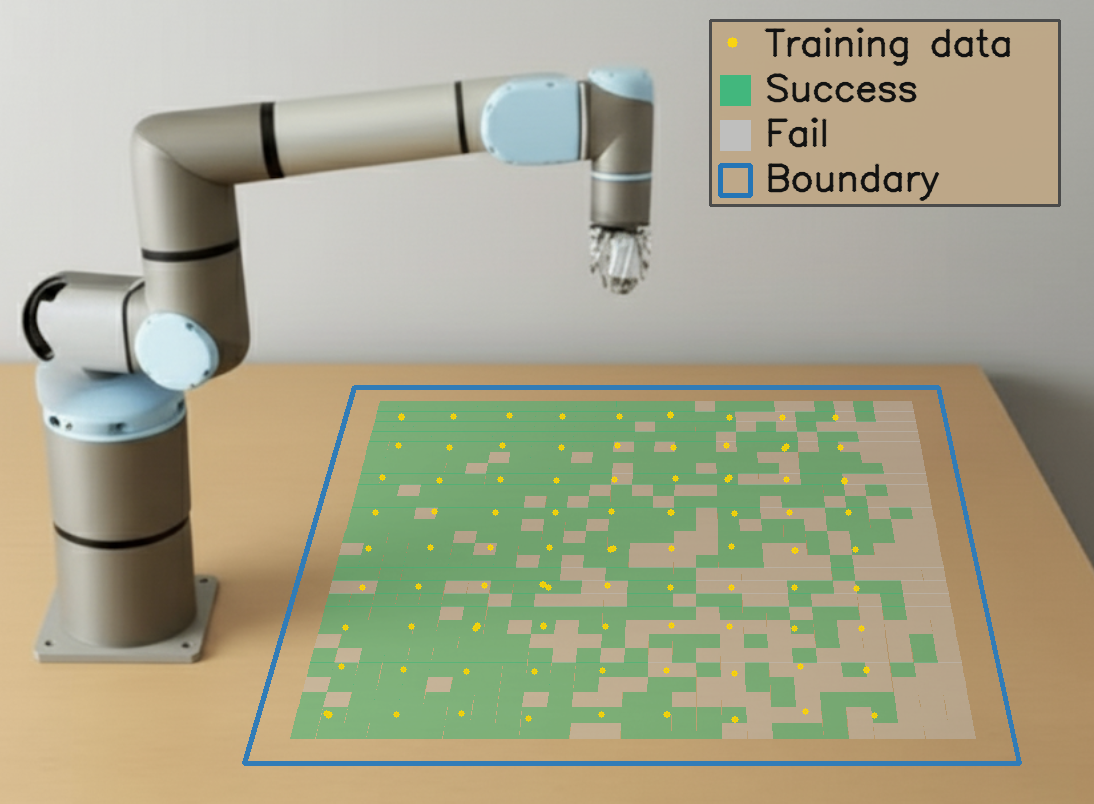}
\end{minipage}\hfill 
\begin{minipage}{0.235\textwidth}
\centering
\includegraphics[width=1\textwidth]{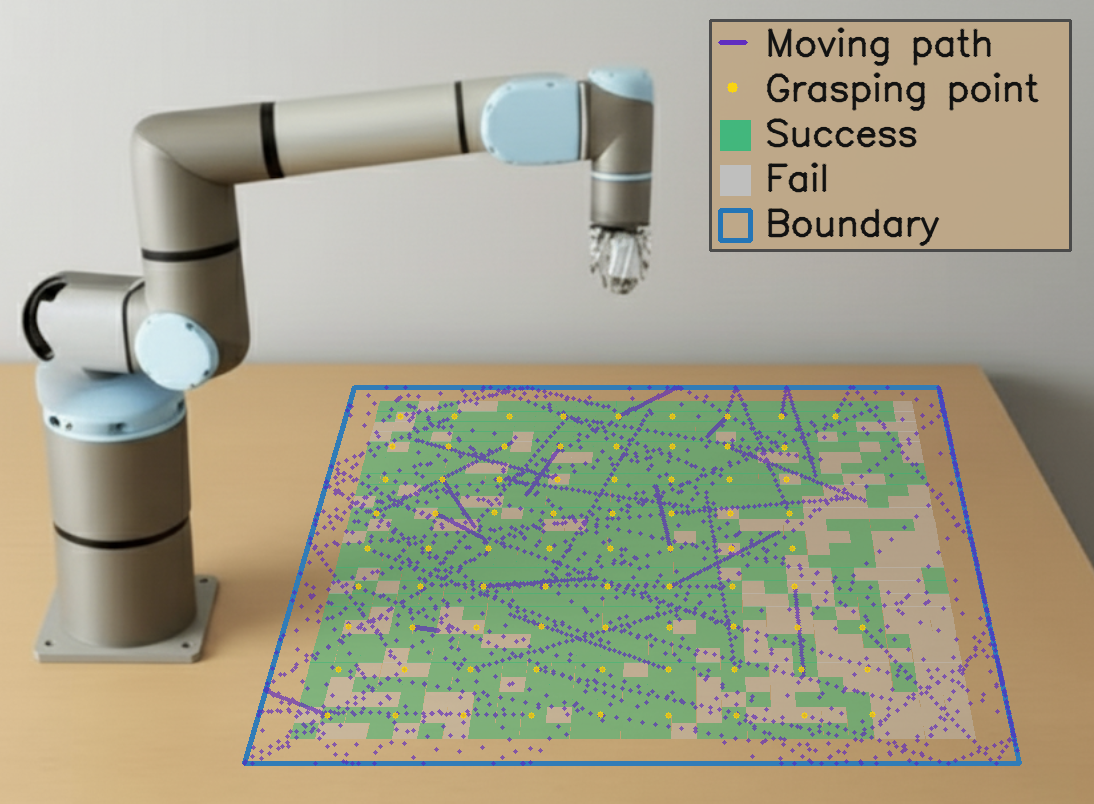}
\end{minipage}\hfill
\caption{
\textbf{Generalization Comparison from Dense Sampling}.
To ensure a fair comparison, we enforce that the grasping point of each MOVE trajectory corresponds to that of a static trajectory.
Despite being exposed to the same set of grasping positions during training, MOVE exhibits significantly better spatial generalization to unseen grasp points (66\% vs. 74\%).
}
\label{fig:pre2}
\end{figure}

\paragraph{Generalization Comparison from Dense Sampling}
While \emph{MOVE} demonstrates strong performance in sparse sampling scenarios, this sampling strategy can be inefficient. Therefore, we validate our method under a dense sampling strategy, which provides better spatial coverage for a fixed dataset size than concentrating on few points. 
We uniformly sample 90 (static) and 85 (\emph{MOVE}) trajectories, ensuring that both datasets have the same total number of timesteps.  
We then evaluate the trained policies across the entire space and visualize the results in Figure~\ref{fig:pre2}.
Although the dense static sampling baseline naturally achieves a higher success rate due to its extensive coverage, \emph{MOVE} still yields substantial improvements from 66\% to 74\%, especially in regions of the state space that are otherwise difficult to learn. 

\begin{figure*}[h]
\centering

\begin{minipage}{0.2\textwidth}
\centering
\includegraphics[width=1\textwidth]{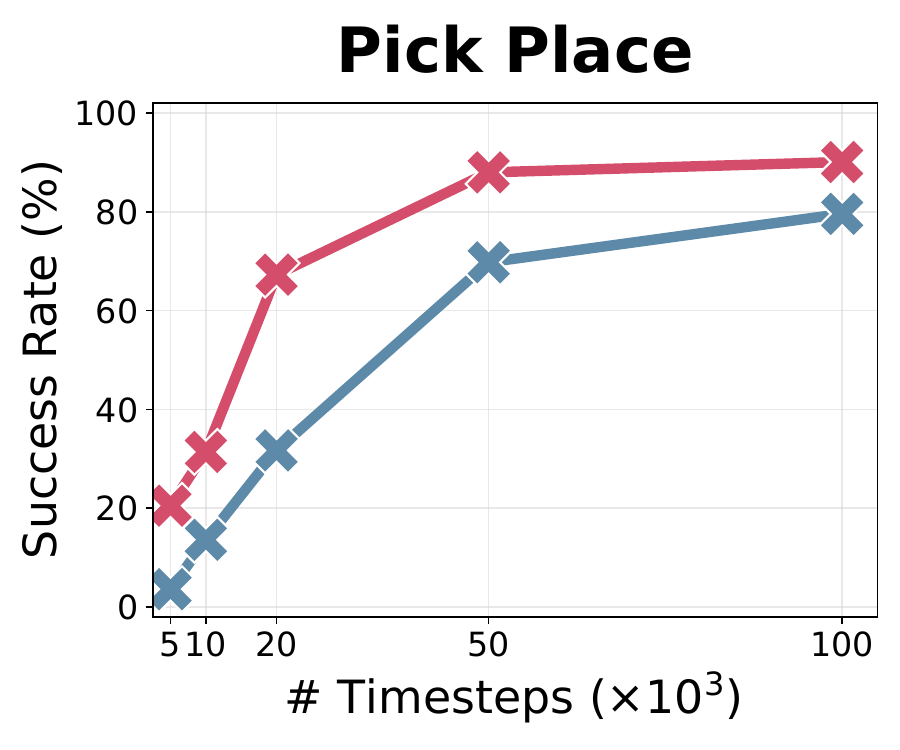}
\end{minipage}\hfill
\begin{minipage}{0.2\textwidth}
\centering
\includegraphics[width=1\textwidth]{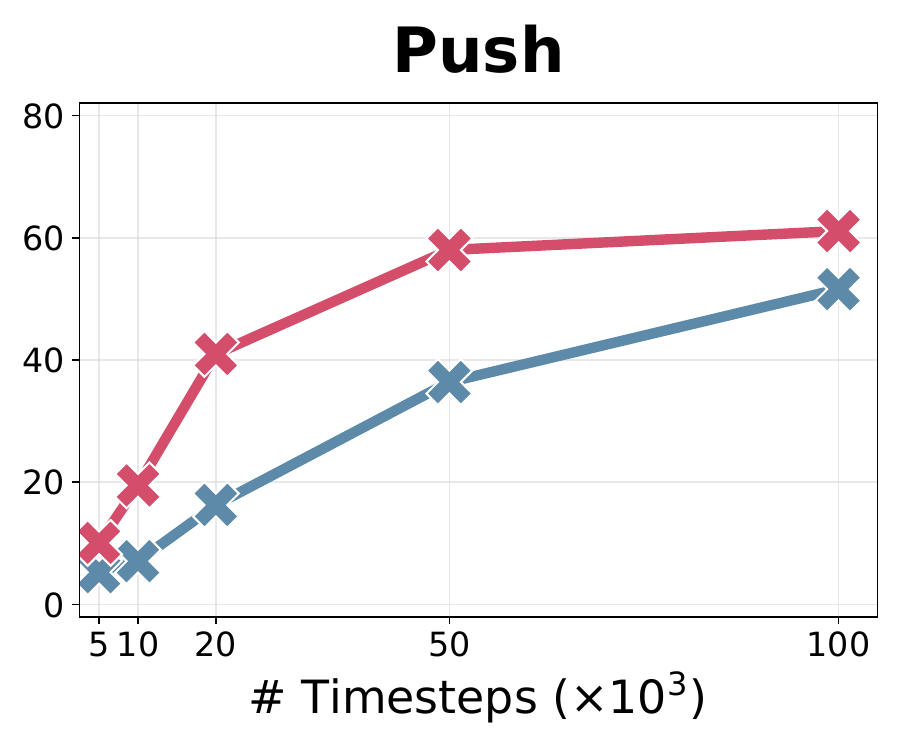}
\end{minipage}\hfill
\begin{minipage}{0.2\textwidth}
\centering
\includegraphics[width=1\textwidth]{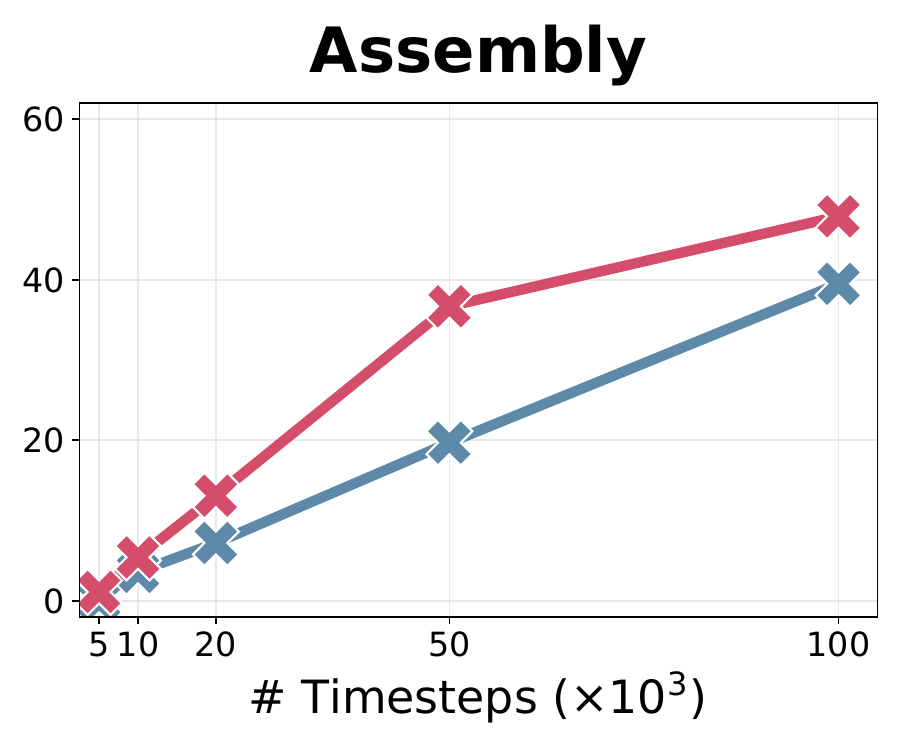}
\end{minipage}\hfill
\begin{minipage}{0.2\textwidth}
\centering
\includegraphics[width=1\textwidth]{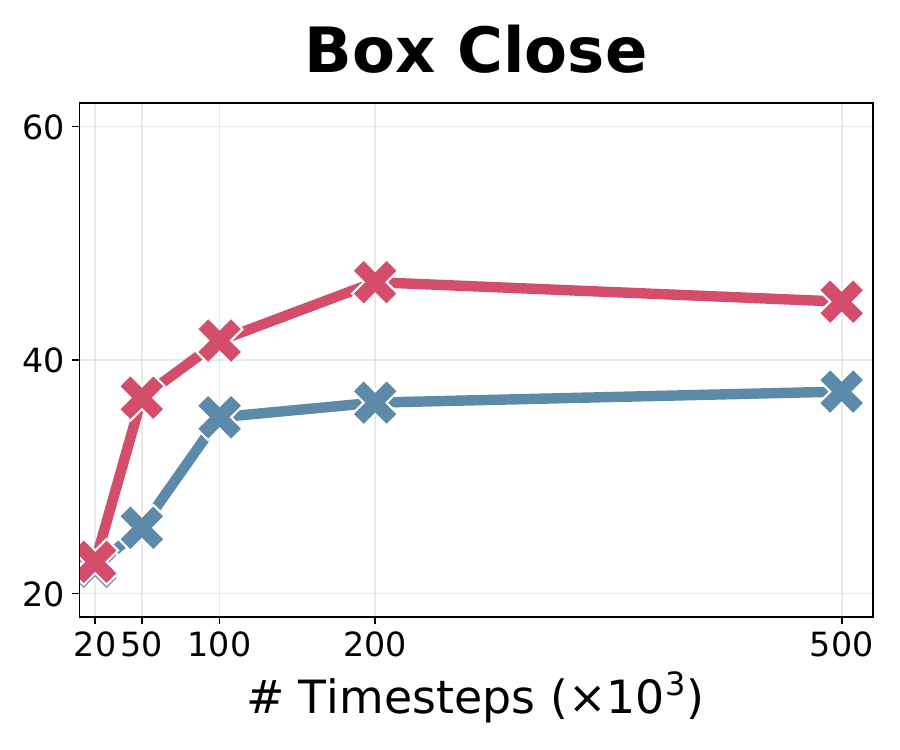}
\end{minipage}\hfill
\begin{minipage}{0.2\textwidth}
\centering
\includegraphics[width=1\textwidth]{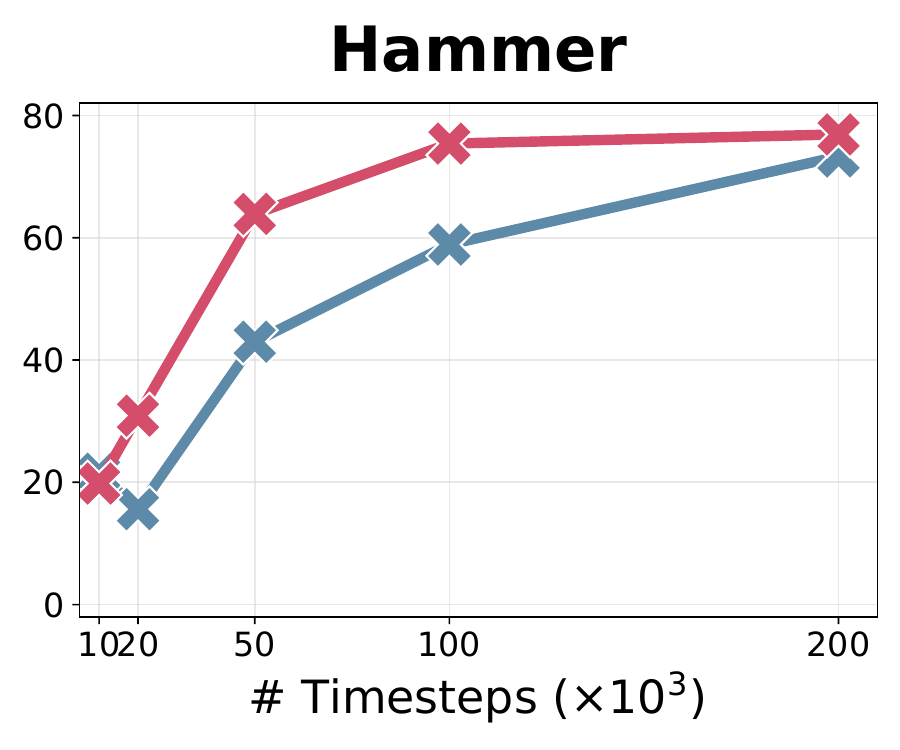}
\end{minipage}\hfill \\

\begin{minipage}{0.2\textwidth}
\centering
\includegraphics[width=1\textwidth]{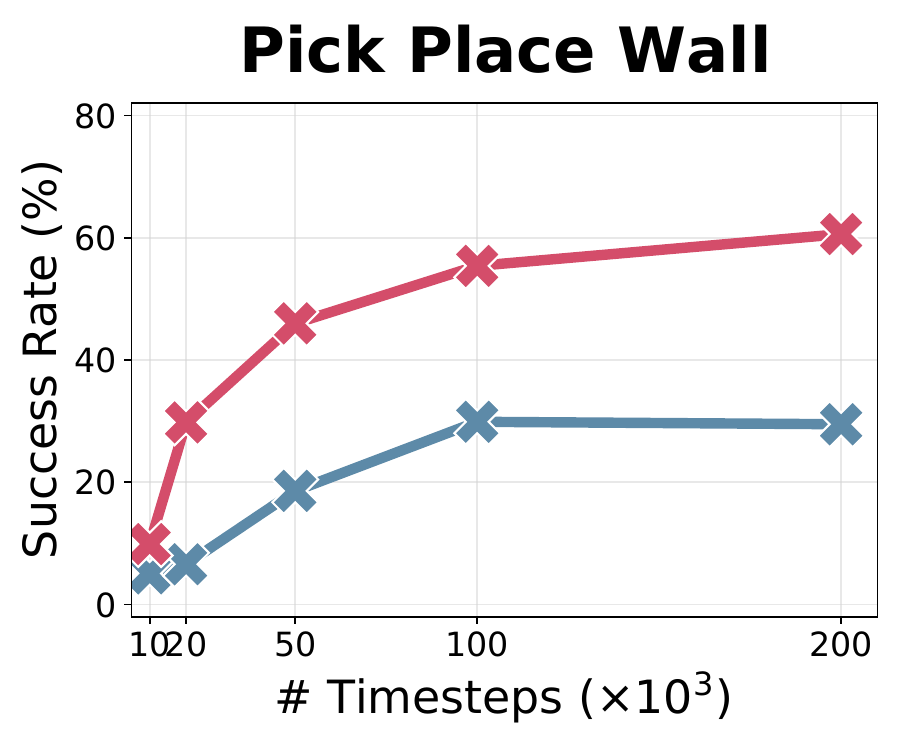}
\end{minipage}\hfill
\begin{minipage}{0.2\textwidth}
\centering
\includegraphics[width=1\textwidth]{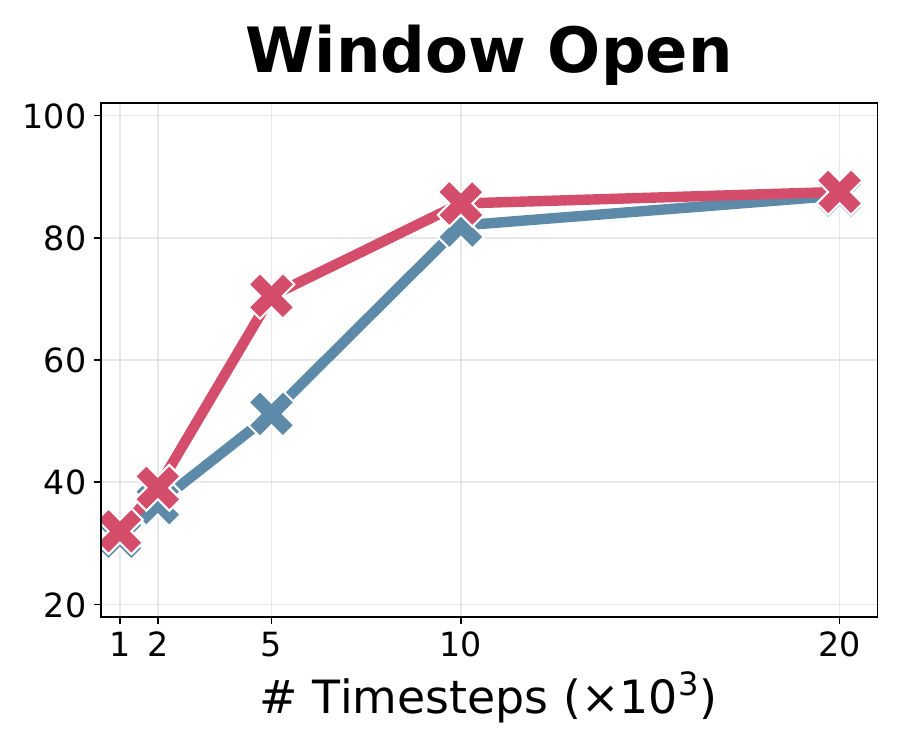}
\end{minipage}\hfill
\begin{minipage}{0.2\textwidth}
\centering
\includegraphics[width=1\textwidth]{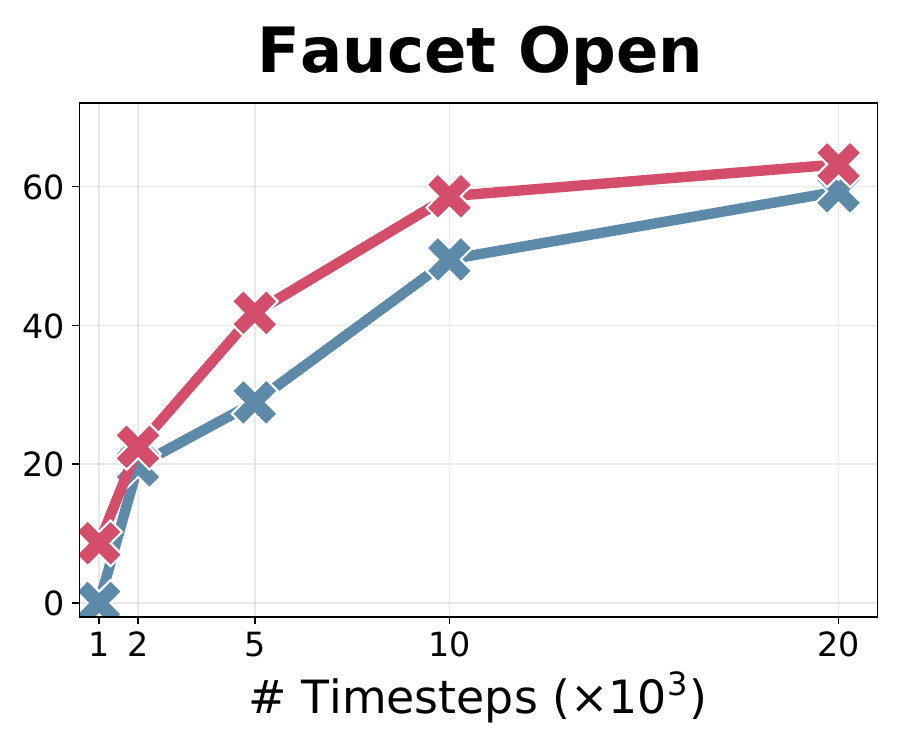}
\end{minipage}\hfill
\begin{minipage}{0.2\textwidth}
\centering
\includegraphics[width=1\textwidth]{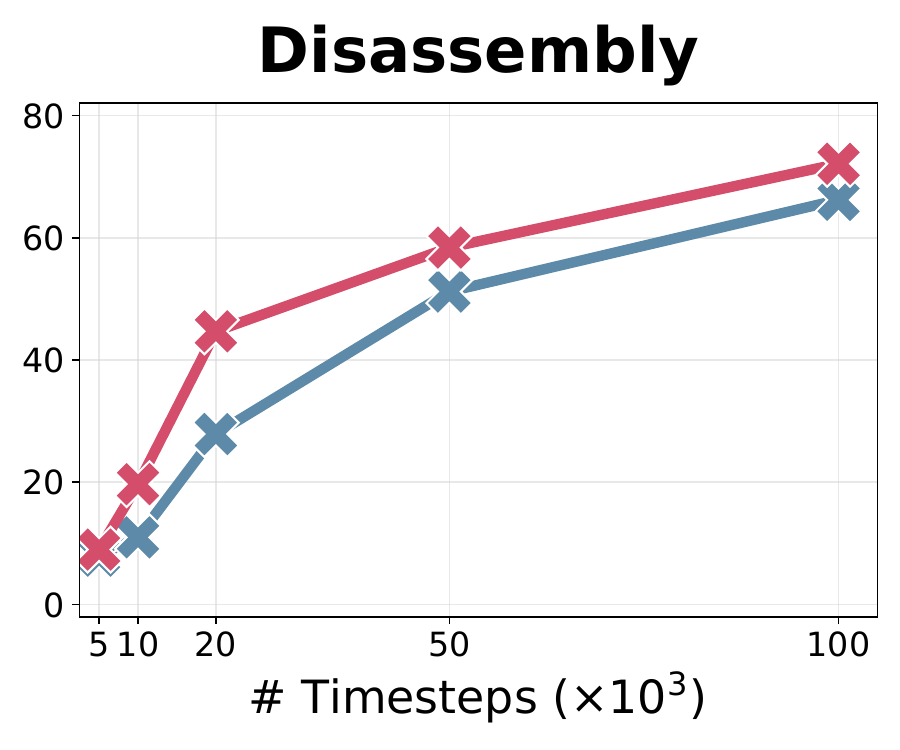}
\end{minipage}\hfill
\begin{minipage}{0.2\textwidth}
\centering
\includegraphics[width=1\textwidth]{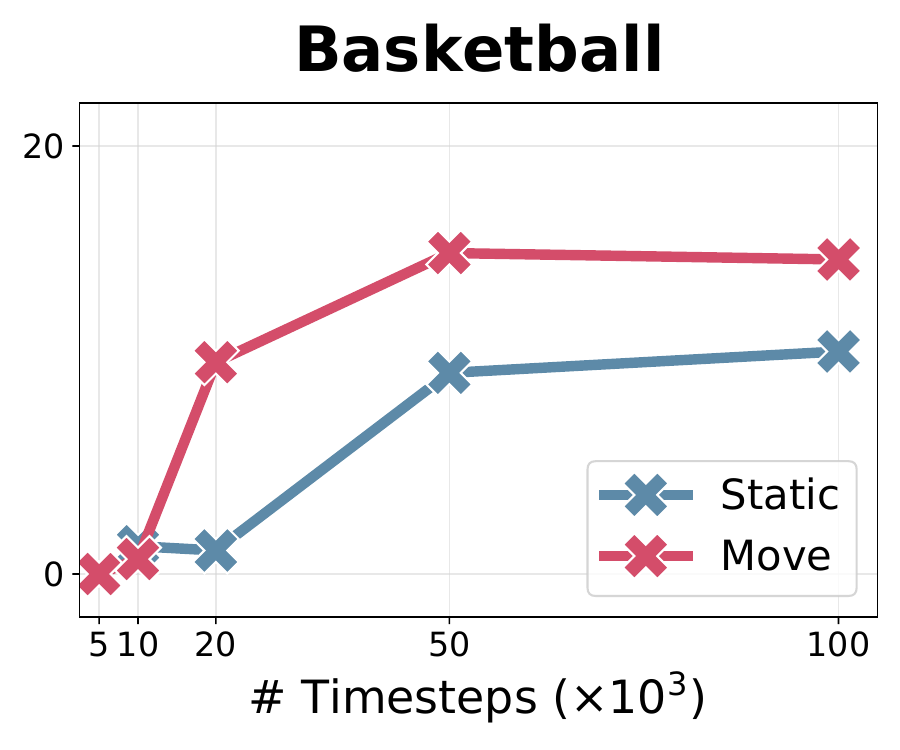}
\end{minipage}\hfill \\
\caption{\textbf{Efficient scaling with demonstrations.} Success rate across 10 simulation tasks. Specifically, the x-axis represents the number of timesteps, where each timestep corresponds to a single robot action, rather than the number of trajectories. 
\emph{MOVE} consistently outperforms the static data collection paradigm at each data scaling point.}
\label{fig:ten-simu}
\end{figure*}


\begin{table*}[t]
\centering
\caption{
\textbf{Main results from the Meta-World simulation environment.}
We \emph{simulate real-world spatial generalization challenges} by testing each time under randomized spatial configurations (object and target positions, camera view points, \textit{et, al}).
We rerun both data collection and training processes with 3 distinct random seeds and report the average success rate.
We ensure all methods use the same total number of timesteps, and thus require nearly the same amount of human effort.
The amount of training data is provided in~\cref{tab:train_data}.}
\label{table:simulation}
\resizebox{1.0\textwidth}{!}{%
\begin{tabular}{c|cccccccccc|cc}
\toprule

\multirow{2}{*}{Method}  & Pick & \multirow{2}{*}{Push} &  \multirow{2}{*}{Assembly} & Box &  \multirow{2}{*}{Hammer}& Pick Place & Window & Faucet & \multirow{2}{*}{Disassembly}   & \multirow{2}{*}{Basketball}   &  \multicolumn{2}{c}{\multirow{2}{*}{Average}}\\
& Place & & & Close & & Wall & Open & Open & & &\\


\midrule

Static & $0.317$ & 0.163 & $0.072$ & $0.351$ & $0.156$ & $0.066$ &  $0.512$ & $0.289$ & $0.279$ & $0.011$ & \multicolumn{2}{c}{0.222{}}\\

ADC~\cite{huang2025adversarialdatacollectionhumancollaborative} & 0.472 & 0.335 & 0.053 & 0.383 & 0.252 & 0.237 & 0.455 & 0.345 & 0.207 & 0.025 & \multicolumn{2}{c}{0.276{}}\\

\textbf{MOVE} &  \textbf{0.673} &  \textbf{0.410} & \textbf{0.131} & \textbf{0.417}& \textbf{0.309} & \textbf{0.298} &  \textbf{0.705} & \textbf{0.418} & \textbf{0.447} & \textbf{0.099} &  \multicolumn{2}{c}{\textbf{0.391} ($\uparrow \mathbf{76.1}\%$)}\\

\bottomrule
\end{tabular}}
\end{table*}

\paragraph{Generalization Comparison from Circle Sampling}

\begin{figure}[h]
\centering

\begin{minipage}{0.235\textwidth}
\centering
\includegraphics[width=1\textwidth]{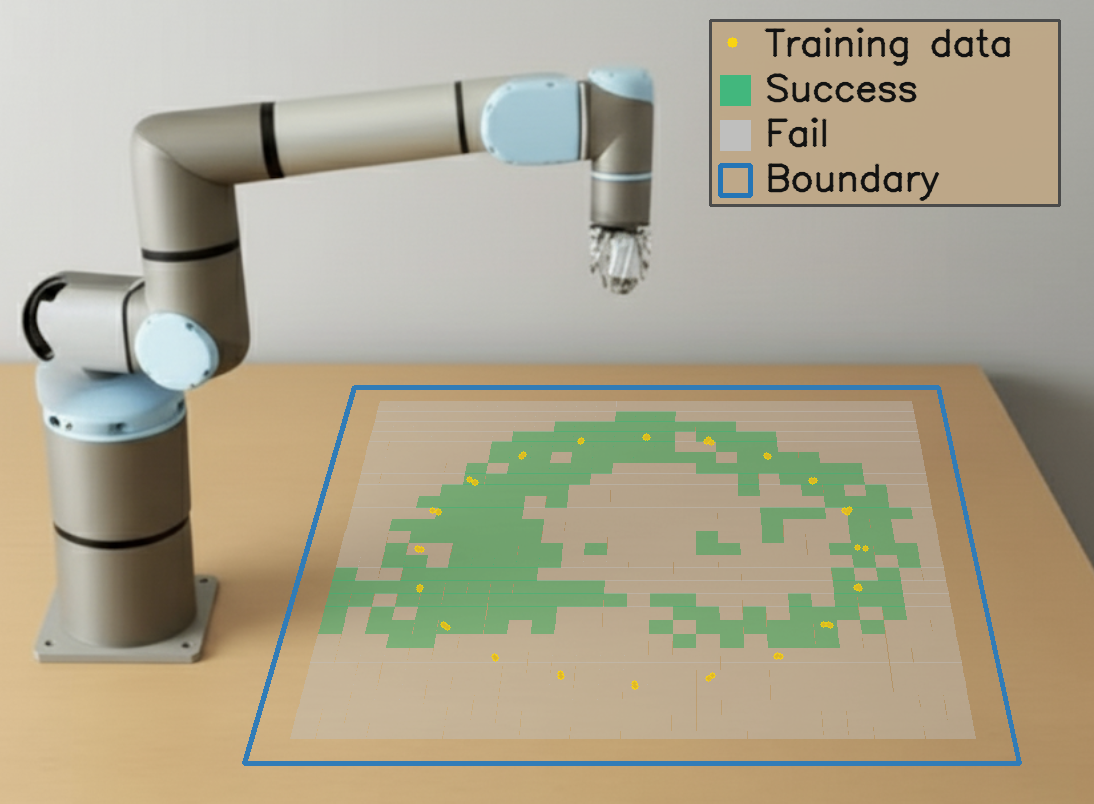}
\end{minipage}\hfill
\begin{minipage}{0.235\textwidth}
\centering
\includegraphics[width=1\textwidth]{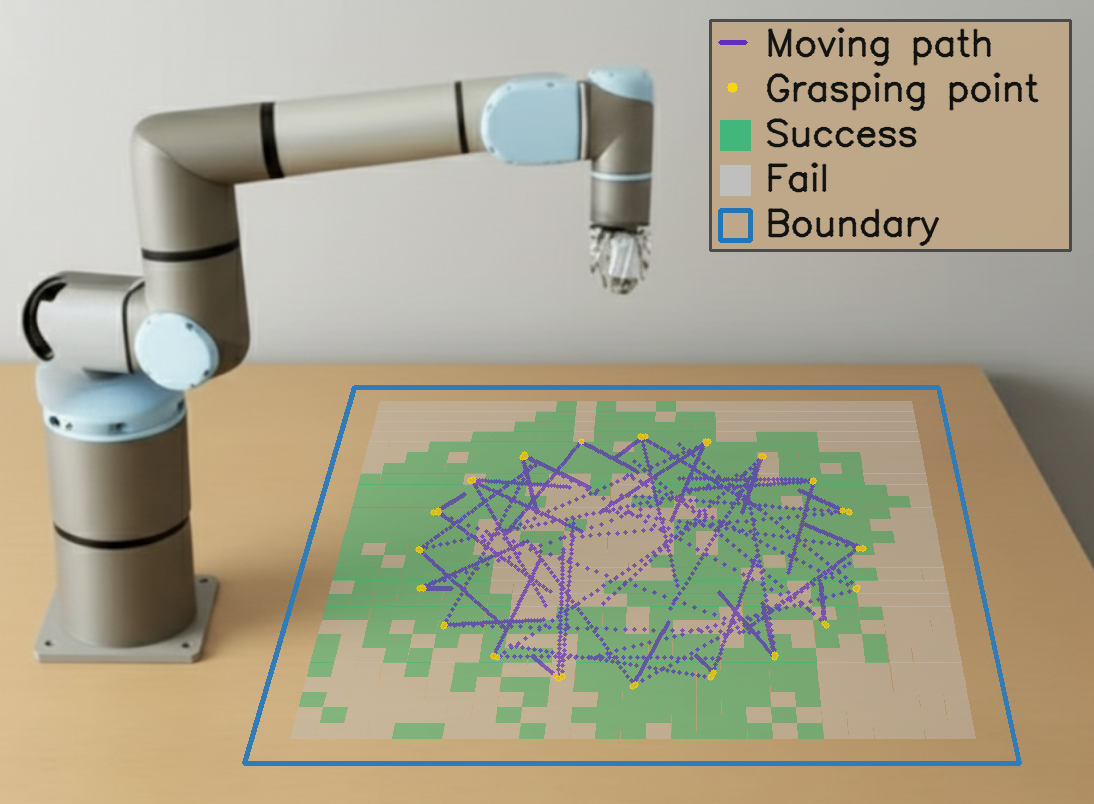}
\end{minipage}\hfill 
\caption{We uniformly sample grasping points evenly distributed on a circle and constrain the object's motion path within this circle for \emph{MOVE}.
While being exposed to the same grasping positions and constrained within the circle, MOVE exhibits significantly better spatial generalization on both the in-circle region (21\% vs. 44\%) and the out-of-circle region (45\% vs. 67\%).}
\label{fig:pre3}
\vspace{-6mm}
\end{figure}

To investigate the influence of data sampling location on \emph{MOVE}'s performance, we sample grasping points evenly distributed on a circle and constrain the object's motion path within this circle. 
We sample 90 (static) and 59 (\emph{MOVE}) trajectories, ensuring that both datasets have the same total number of timesteps.  
We evaluate both policies across the entire space and visualize the results in Figure~\ref{fig:pre3}.
Surprisingly, \emph{MOVE} not only dominates within the sampling circle but also significantly outperforms the baseline in the outer regions.
Specifically, \emph{MOVE} achieves a success rate of 43.7\% and outperforms the static policy's 21.3\% across the in-circle space; \emph{MOVE} also achieves a success rate of 67.4\% while the static policy achieves 44.6\% across the out-of-circle space.

\subsection{Experiment Setup}

The following experiments are designed to comprehensively evaluate the effectiveness of \emph{MOVE}, and compare this method with the traditional static paradigm. We conduct experiments in both simulation and real-world environments.

\paragraph{Tasks and Environments} In simulation environments, we leverage the Meta-World benchmark~\cite{yu2021metaworldbenchmarkevaluationmultitask} which comprises a series of robotic manipulation tasks including grasping, pushing and placement. 
For each simulation environment, we implement random initialization of the object, target, and camera over wide ranges by modifying the simulator code.
We select a specific number of environments from each of three difficulty levels as representative examples due to time-costly code modification. 

For real-world validation, we use a canonical pick-and-place task. An Agilex PIPER arm is tasked with grasping an orange from a variable initial position and placing it onto a plastic tray. The operational workspace is configured as a semicircle, covering an $0.5\times\pi\times60\text{ cm}\times60\text{ cm}\approx 5655\text{cm}^2$ area, which encompasses the majority of the end-effector’s reachable space. The training data set was constructed by randomly sampling 20 pairs of positions, where each pair specifies one orange location and one plate location. More details  are provided in the appendix~\ref{app:real}.

\paragraph{Baselines} 
To demonstrate the benefits of our approach, we compare \emph{MOVE} against the conventional static data collection paradigm.
Furthermore, we compare against the ADC method~\cite{huang2025adversarialdatacollectionhumancollaborative}, which periodically resets the object's position to a new random location during a single data collection trajectory to encourage policy diversity. For all methods, we use the same training and evaluation protocols.

\paragraph{Expert Data Collection}  In simulation, expert demonstrations are generated using the scripted policies provided by Meta-World. In the real world, we collect human demonstrations by teleoperating the robot arm using the Pika gripper. A key consideration is that trajectories collected under dynamic conditions are often longer than static ones (see~\cref{table:length}). To ensure a fair comparison of data efficiency, we define the dataset size by the total number of environment interaction steps rather than the number of trajectories.


\subsection{Results in Simulated Environments}

We first evaluate \emph{MOVE} on Meta-World. Performance in 10 simulation tasks with varying demonstration sizes is visualized in Figure~\ref{fig:ten-simu}, and detailed average success rates are presented in Table~\ref{table:simulation}. 
The results illustrate that while both \emph{MOVE} consistently and remarkably outperforms the static data collection paradigm across all tasks and dataset sizes. Specifically, our approach improves the success rate by 76.1\% with equal data. In the Pick-Place-Wall task, \emph{MOVE} achieves comparable performance on the 20k-sized dynamic dataset to the 100k-sized static dataset, demonstrating a data efficiency of up to 5x.

\subsection{Results in Real Environments}

In this subsection, we transition the experiments to a real-world environment to validate the effectiveness of our method. For model training, we collected datasets comprising 35k and 75k timsteps, respectively. 
For evaluation, we test the learned policy on a grid of 30 initial object positions that were unseen during training, sampled from a $40\text{cm }\times 80\text{ cm}$ workspace located within the semicircle. The detailed results are presented in Table~\ref{table:real}. Notably, when the dataset size is 35k, the performance of \emph{MOVE} is nearly comparable to that of the static data collection method using a 75k dataset.

\begin{table}[h]
\centering
\caption{\textbf{Main results from the real environment.} We report the success rate and normalized score across different dataset sizes. Following prior work~\cite{lin2025datascalinglawsimitation}, we divide the task into three steps: approaching the correct grasping pose, picking up the orange, and putting it on the plate. Each successful step is scored 1 point. We report a normalized score, defined as $\text{Normalized score} =\frac{\text{Total test score}}{3\times \text{Number of steps}}$, with a maximum value of 1.}
\label{table:real}
\begin{tabular}{ll|cc}
\toprule
\# Timesteps & Method & Success Rate & Normalized Score\\ 
\midrule
\multirowcell{2}{35k} & static &  3.3\% & 0.055  \\
 & \textbf{MOVE} &  \cellcolor{gray!20}\textbf{23.3\%} & \cellcolor{gray!20}\textbf{0.311} \\
 \midrule
\multirowcell{2}{75k} & static &  23.3\% & 0.389 \\
 & \textbf{MOVE}&  \cellcolor{gray!20}\textbf{36.7\%} & \cellcolor{gray!20}\textbf{0.522}\\
\bottomrule
\end{tabular}
\vspace{-5mm}
\end{table}

\subsection{Ablation Study}
\label{sec:abla}

To validate our design choices, we conduct ablation studies on two representative tasks: Pick-Place and Assembly.

\paragraph{The Impact of Dynamic Dimension Combination}  
We hypothesize that combining multiple dynamic spatial dimensions, such as pickup object position, target object position, and camera viewpoint, within a single trajectory enriches spatial information and is critical for learning a robust and generalizable policy.  
Therefore, \emph{MOVE} employs combinations of dynamic dimensions during data collection to maximize spatial diversity.

\begin{figure}[h]
\centering

\begin{minipage}{0.24\textwidth}
\centering
\includegraphics[width=1\textwidth]{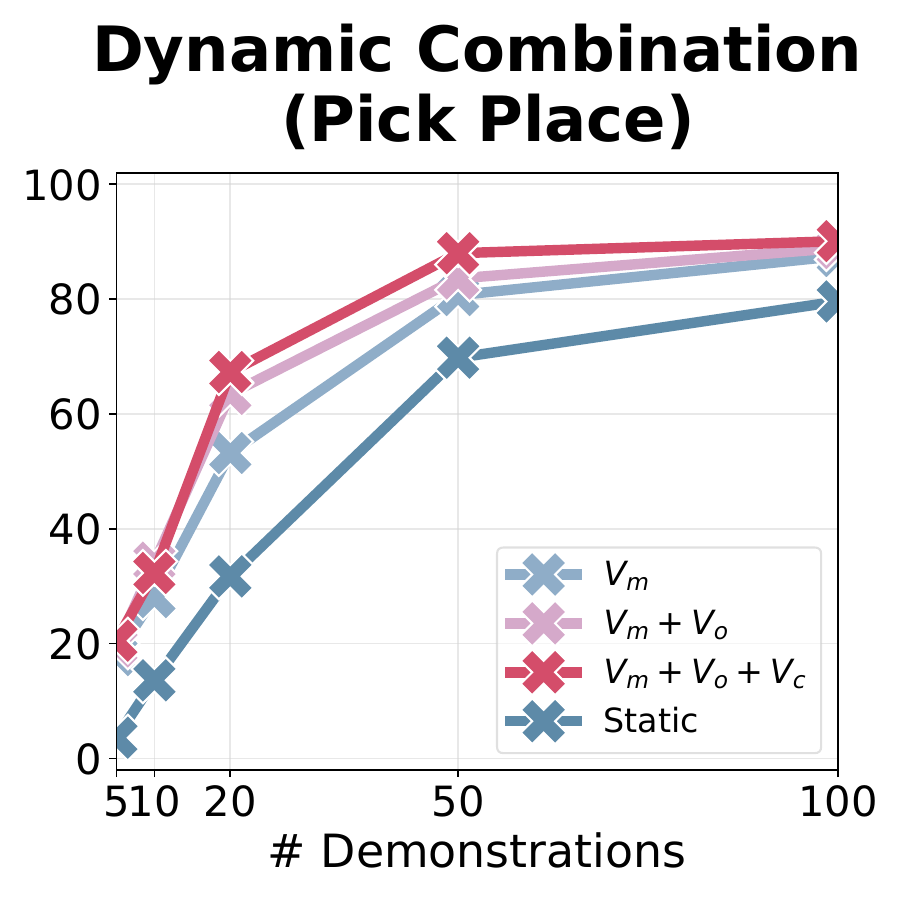}
\end{minipage}\hfill
\begin{minipage}{0.24\textwidth}
\centering
\includegraphics[width=1\textwidth]{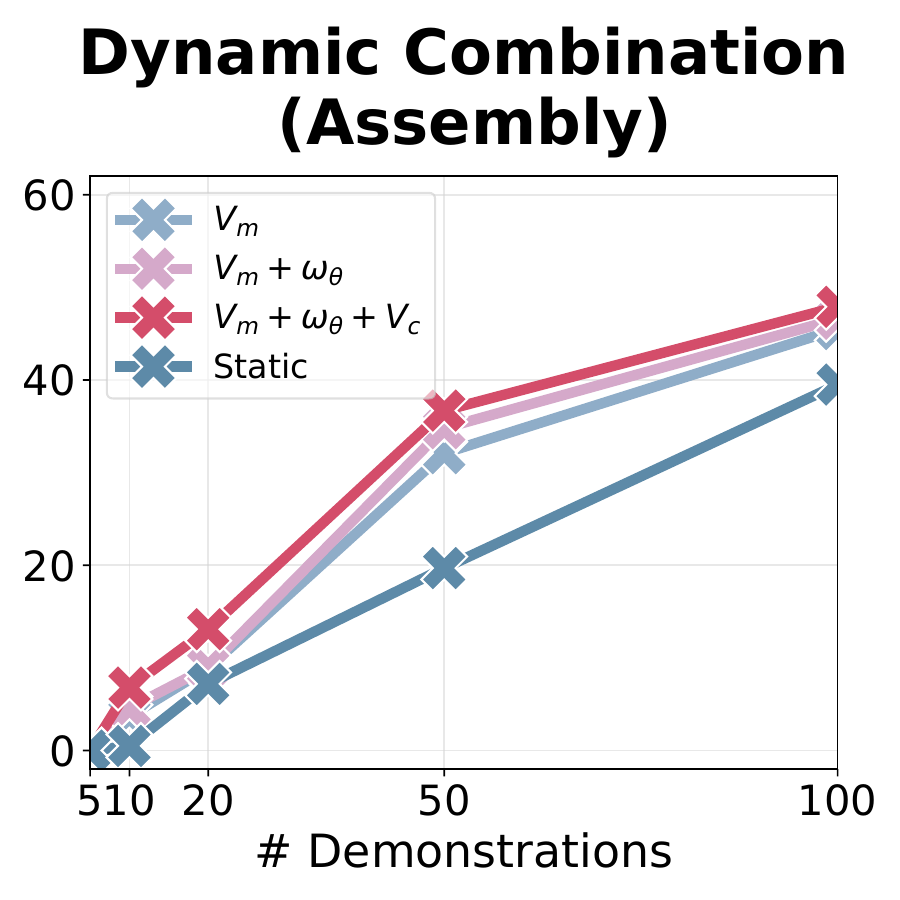}
\end{minipage}\hfill 
\caption{
\textbf{The Impact of Dynamic Dimension Combination.}  
$V_m$ refers to \emph{MOVE} applied only to the pickup object's motion.  
$+V_o$ indicates the addition of target object motion,  $+V_c$ further incorporates dynamic camera movement,
$+\omega_\theta$ denotes dynamic object rotation.
}
\label{fig:abla-comb}
\end{figure}

To validate this, we follow the same setup as in the main experiments, but selectively apply \emph{MOVE} to individual spatial dimensions while keeping the others static.  
The results, shown in Figure~\ref{fig:abla-comb}, highlight the impact of combinations of dynamic dimensions.  
For example, in the Pick-Place task, we start with object translation as the initial dynamic component, then incrementally add target object translation, and finally incorporate camera motion.  
The results demonstrate that as more dynamic dimensions are introduced in \emph{MOVE}, the success rate consistently improves.

\begin{figure}[h]
\centering

\begin{minipage}{0.24\textwidth}
\centering
\includegraphics[width=1\textwidth]{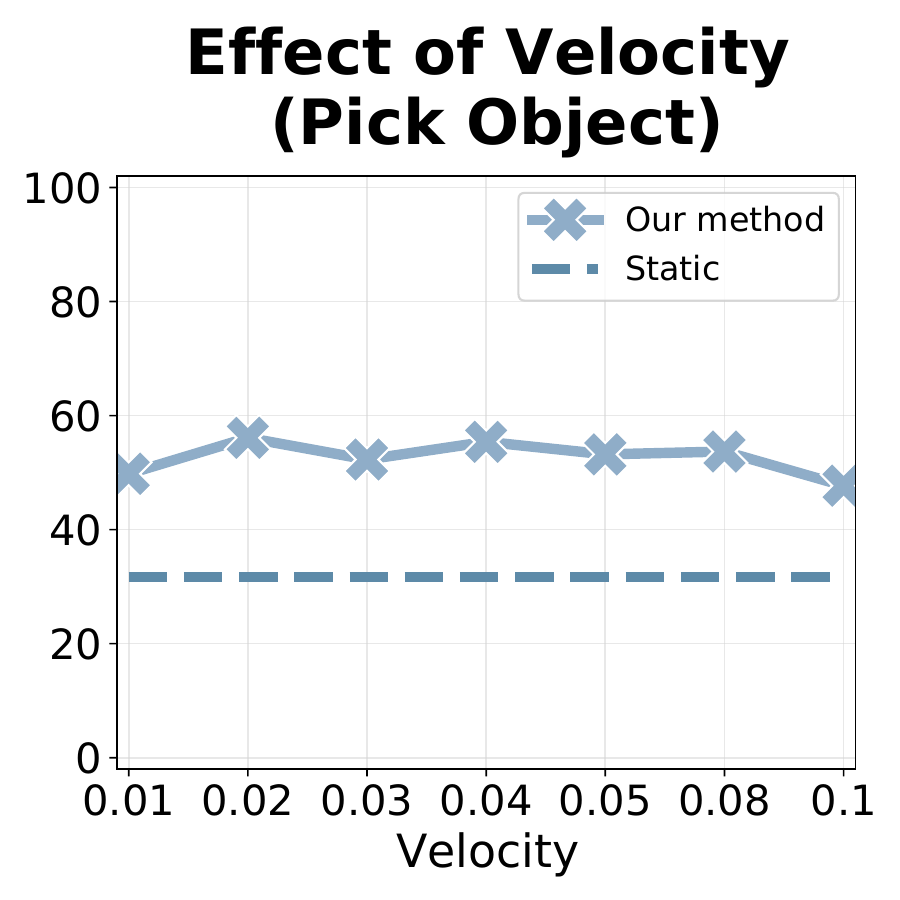}
\end{minipage}\hfill
\begin{minipage}{0.24\textwidth}
\centering
\includegraphics[width=1\textwidth]{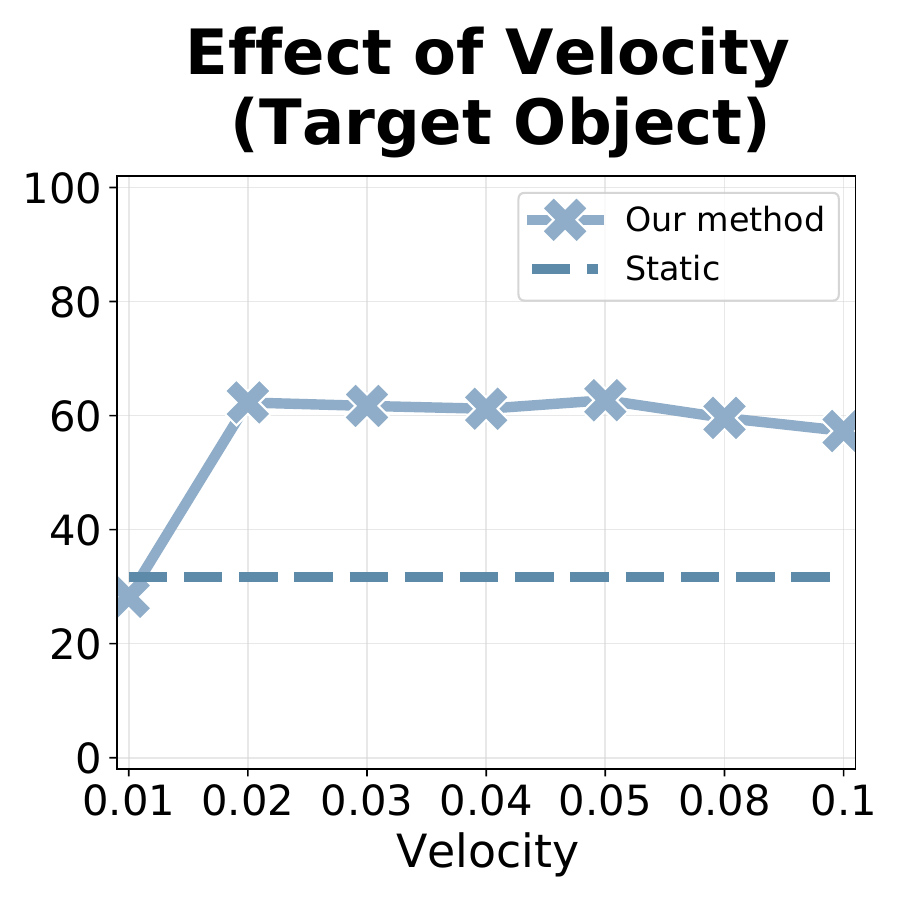}
\end{minipage}\hfill \\
\caption{The impact of $v_\text{max}$ on success rate for the pickup object (left) and the target object (right).}
\label{fig:abla-vel}
\vspace{-2mm}
\end{figure}

\paragraph{The Impact of Augmentation Hyperparameters}  
The maximum velocities, $v_\text{max}$ (translation), $\omega_\text{max}$ (rotation), and $u_\text{max}$ (camera motion), serve as hyperparameters that control the intensity of spatial augmentation (see~\cref{method:dyna}).  
To evaluate their impact, we select $v_\text{max}$ at varying speed levels and present the results in Figure~\ref{fig:abla-vel}.  
The results demonstrate that our method exhibits a degree of robustness to these speed variations.

\vspace{-2mm}

\section{Conclusion and Limitation}

In this paper, we introduce a simple yet effective data collection paradigm that significantly enhances spatial generalization in robotic manipulation and alleviate the challenge of data scarcity. By incorporating dynamic spatial configurations into demonstrations, \emph{MOVE} provide much richer spatial information in each trajectory. 
This is evidenced by \emph{MOVE}'s consistent improvements over static data collection across varying dataset scales. 
While our study demonstrates promising results, it also has several limitations.
For instance, it remains to be explored how \emph{MOVE} can be applied in real-world environments to spatial variations such as changing camera viewpoints, as well as to more complex manipulation tasks like folding garments.
We expect this limitation to be mitigated by developing mechanical devices that can adjust camera viewpoints and introduce perturbations such as translations, shuffling, and rotations to garments.
We leave these directions to future work.



\bibliographystyle{IEEEtran}
\bibliography{custom.bib}

@IEEEtranBSTCTL{IEEEexample:BSTcontrol,
  CTLuse_forced_etal       = "yes",
  CTLname_max_names        = "3",
  CTLnames_show_etal    = "8"
}

@misc{brohan2023rt2visionlanguageactionmodelstransfer,
      title={RT-2: Vision-Language-Action Models Transfer Web Knowledge to Robotic Control}, 
      author={Anthony Brohan and Noah Brown and Justice Carbajal and Yevgen Chebotar and Xi Chen and Krzysztof Choromanski and Tianli Ding and Danny Driess and Avinava Dubey and Chelsea Finn and Pete Florence and Chuyuan Fu and Montse Gonzalez Arenas and Keerthana Gopalakrishnan and Kehang Han and Karol Hausman and Alexander Herzog and Jasmine Hsu and Brian Ichter and Alex Irpan and Nikhil Joshi and Ryan Julian and Dmitry Kalashnikov and Yuheng Kuang and Isabel Leal and Lisa Lee and Tsang-Wei Edward Lee and Sergey Levine and Yao Lu and Henryk Michalewski and Igor Mordatch and Karl Pertsch and Kanishka Rao and Krista Reymann and Michael Ryoo and Grecia Salazar and Pannag Sanketi and Pierre Sermanet and Jaspiar Singh and Anikait Singh and Radu Soricut and Huong Tran and Vincent Vanhoucke and Quan Vuong and Ayzaan Wahid and Stefan Welker and Paul Wohlhart and Jialin Wu and Fei Xia and Ted Xiao and Peng Xu and Sichun Xu and Tianhe Yu and Brianna Zitkovich},
      year={2023},
      eprint={2307.15818},
      archivePrefix={arXiv},
      primaryClass={cs.RO},
      note = {\url{https://arxiv.org/abs/2307.15818}}, 
}

@misc{cheang2024gr2generativevideolanguageactionmodel,
      title={GR-2: A Generative Video-Language-Action Model with Web-Scale Knowledge for Robot Manipulation}, 
      author={Chi-Lam Cheang and Guangzeng Chen and Ya Jing and Tao Kong and Hang Li and Yifeng Li and Yuxiao Liu and Hongtao Wu and Jiafeng Xu and Yichu Yang and Hanbo Zhang and Minzhao Zhu},
      year={2024},
      eprint={2410.06158},
      archivePrefix={arXiv},
      primaryClass={cs.RO},
      note = {\url{https://arxiv.org/abs/2410.06158}}, 
}

@inproceedings{octomodelteam2024octoopensourcegeneralistrobot,
    title={Octo: An Open-Source Generalist Robot Policy},
    author = {{Octo Model Team} and Dibya Ghosh and Homer Walke and Karl Pertsch and Kevin Black and Oier Mees and Sudeep Dasari and Joey Hejna and Charles Xu and Jianlan Luo and Tobias Kreiman and {You Liang} Tan and Lawrence Yunliang Chen and Pannag Sanketi and Quan Vuong and Ted Xiao and Dorsa Sadigh and Chelsea Finn and Sergey Levine},
    booktitle = {Proceedings of Robotics: Science and Systems},
    year = {2024},
}

@misc{kim2024openvlaopensourcevisionlanguageactionmodel,
      title={OpenVLA: An Open-Source Vision-Language-Action Model}, 
      author={Moo Jin Kim and Karl Pertsch and Siddharth Karamcheti and Ted Xiao and Ashwin Balakrishna and Suraj Nair and Rafael Rafailov and Ethan Foster and Grace Lam and Pannag Sanketi and Quan Vuong and Thomas Kollar and Benjamin Burchfiel and Russ Tedrake and Dorsa Sadigh and Sergey Levine and Percy Liang and Chelsea Finn},
      year={2024},
      eprint={2406.09246},
      archivePrefix={arXiv},
      primaryClass={cs.RO},
      note = {\url{https://arxiv.org/abs/2406.09246}}, 
}

@inproceedings{liu2025rdt1bdiffusionfoundationmodel,
title={{RDT}-1B: a Diffusion Foundation Model for Bimanual Manipulation},
author={Songming Liu and Lingxuan Wu and Bangguo Li and Hengkai Tan and Huayu Chen and Zhengyi Wang and Ke Xu and Hang Su and Jun Zhu},
booktitle={The Thirteenth International Conference on Learning Representations},
year={2025},
url={https://openreview.net/forum?id=yAzN4tz7oI}

}

@article{chi2024diffusionpolicy,
	author = {Cheng Chi and Zhenjia Xu and Siyuan Feng and Eric Cousineau and Yilun Du and Benjamin Burchfiel and Russ Tedrake and Shuran Song},
	title ={Diffusion Policy: Visuomotor Policy Learning via Action Diffusion},
	journal = {Robotics: Science and Systems},
	year = {2023},
}

@misc{embodimentcollaboration2025openxembodimentroboticlearning,
      title={Open X-Embodiment: Robotic Learning Datasets and RT-X Models}, 
      author={Embodiment Collaboration and Abby O'Neill and Abdul Rehman and Abhinav Gupta and Abhiram Maddukuri and Abhishek Gupta and Abhishek Padalkar and Abraham Lee and Acorn Pooley and Agrim Gupta and Ajay Mandlekar and Ajinkya Jain and Albert Tung and Alex Bewley and Alex Herzog and Alex Irpan and Alexander Khazatsky and Anant Rai and Anchit Gupta and Andrew Wang and Andrey Kolobov and Anikait Singh and Animesh Garg and Aniruddha Kembhavi and Annie Xie and Anthony Brohan and Antonin Raffin and Archit Sharma and Arefeh Yavary and Arhan Jain and Ashwin Balakrishna and Ayzaan Wahid and Ben Burgess-Limerick and Beomjoon Kim and Bernhard Schölkopf and Blake Wulfe and Brian Ichter and Cewu Lu and Charles Xu and Charlotte Le and Chelsea Finn and Chen Wang and Chenfeng Xu and Cheng Chi and Chenguang Huang and Christine Chan and Christopher Agia and Chuer Pan and Chuyuan Fu and Coline Devin and Danfei Xu and Daniel Morton and Danny Driess and Daphne Chen and Deepak Pathak and Dhruv Shah and Dieter Büchler and Dinesh Jayaraman and Dmitry Kalashnikov and Dorsa Sadigh and Edward Johns and Ethan Foster and Fangchen Liu and Federico Ceola and Fei Xia and Feiyu Zhao and Felipe Vieira Frujeri and Freek Stulp and Gaoyue Zhou and Gaurav S. Sukhatme and Gautam Salhotra and Ge Yan and Gilbert Feng and Giulio Schiavi and Glen Berseth and Gregory Kahn and Guangwen Yang and Guanzhi Wang and Hao Su and Hao-Shu Fang and Haochen Shi and Henghui Bao and Heni Ben Amor and Henrik I Christensen and Hiroki Furuta and Homanga Bharadhwaj and Homer Walke and Hongjie Fang and Huy Ha and Igor Mordatch and Ilija Radosavovic and Isabel Leal and Jacky Liang and Jad Abou-Chakra and Jaehyung Kim and Jaimyn Drake and Jan Peters and Jan Schneider and Jasmine Hsu and Jay Vakil and Jeannette Bohg and Jeffrey Bingham and Jeffrey Wu and Jensen Gao and Jiaheng Hu and Jiajun Wu and Jialin Wu and Jiankai Sun and Jianlan Luo and Jiayuan Gu and Jie Tan and Jihoon Oh and Jimmy Wu and Jingpei Lu and Jingyun Yang and Jitendra Malik and João Silvério and Joey Hejna and Jonathan Booher and Jonathan Tompson and Jonathan Yang and Jordi Salvador and Joseph J. Lim and Junhyek Han and Kaiyuan Wang and Kanishka Rao and Karl Pertsch and Karol Hausman and Keegan Go and Keerthana Gopalakrishnan and Ken Goldberg and Kendra Byrne and Kenneth Oslund and Kento Kawaharazuka and Kevin Black and Kevin Lin and Kevin Zhang and Kiana Ehsani and Kiran Lekkala and Kirsty Ellis and Krishan Rana and Krishnan Srinivasan and Kuan Fang and Kunal Pratap Singh and Kuo-Hao Zeng and Kyle Hatch and Kyle Hsu and Laurent Itti and Lawrence Yunliang Chen and Lerrel Pinto and Li Fei-Fei and Liam Tan and Linxi "Jim" Fan and Lionel Ott and Lisa Lee and Luca Weihs and Magnum Chen and Marion Lepert and Marius Memmel and Masayoshi Tomizuka and Masha Itkina and Mateo Guaman Castro and Max Spero and Maximilian Du and Michael Ahn and Michael C. Yip and Mingtong Zhang and Mingyu Ding and Minho Heo and Mohan Kumar Srirama and Mohit Sharma and Moo Jin Kim and Muhammad Zubair Irshad and Naoaki Kanazawa and Nicklas Hansen and Nicolas Heess and Nikhil J Joshi and Niko Suenderhauf and Ning Liu and Norman Di Palo and Nur Muhammad Mahi Shafiullah and Oier Mees and Oliver Kroemer and Osbert Bastani and Pannag R Sanketi and Patrick "Tree" Miller and Patrick Yin and Paul Wohlhart and Peng Xu and Peter David Fagan and Peter Mitrano and Pierre Sermanet and Pieter Abbeel and Priya Sundaresan and Qiuyu Chen and Quan Vuong and Rafael Rafailov and Ran Tian and Ria Doshi and Roberto Martín-Martín and Rohan Baijal and Rosario Scalise and Rose Hendrix and Roy Lin and Runjia Qian and Ruohan Zhang and Russell Mendonca and Rutav Shah and Ryan Hoque and Ryan Julian and Samuel Bustamante and Sean Kirmani and Sergey Levine and Shan Lin and Sherry Moore and Shikhar Bahl and Shivin Dass and Shubham Sonawani and Shubham Tulsiani and Shuran Song and Sichun Xu and Siddhant Haldar and Siddharth Karamcheti and Simeon Adebola and Simon Guist and Soroush Nasiriany and Stefan Schaal and Stefan Welker and Stephen Tian and Subramanian Ramamoorthy and Sudeep Dasari and Suneel Belkhale and Sungjae Park and Suraj Nair and Suvir Mirchandani and Takayuki Osa and Tanmay Gupta and Tatsuya Harada and Tatsuya Matsushima and Ted Xiao and Thomas Kollar and Tianhe Yu and Tianli Ding and Todor Davchev and Tony Z. Zhao and Travis Armstrong and Trevor Darrell and Trinity Chung and Vidhi Jain and Vikash Kumar and Vincent Vanhoucke and Vitor Guizilini and Wei Zhan and Wenxuan Zhou and Wolfram Burgard and Xi Chen and Xiangyu Chen and Xiaolong Wang and Xinghao Zhu and Xinyang Geng and Xiyuan Liu and Xu Liangwei and Xuanlin Li and Yansong Pang and Yao Lu and Yecheng Jason Ma and Yejin Kim and Yevgen Chebotar and Yifan Zhou and Yifeng Zhu and Yilin Wu and Ying Xu and Yixuan Wang and Yonatan Bisk and Yongqiang Dou and Yoonyoung Cho and Youngwoon Lee and Yuchen Cui and Yue Cao and Yueh-Hua Wu and Yujin Tang and Yuke Zhu and Yunchu Zhang and Yunfan Jiang and Yunshuang Li and Yunzhu Li and Yusuke Iwasawa and Yutaka Matsuo and Zehan Ma and Zhuo Xu and Zichen Jeff Cui and Zichen Zhang and Zipeng Fu and Zipeng Lin},
      year={2025},
      eprint={2310.08864},
      archivePrefix={arXiv},
      primaryClass={cs.RO},
      note = {\url{https://arxiv.org/abs/2310.08864}}, 
}

@misc{xue2025demogensyntheticdemonstrationgeneration,
      title={DemoGen: Synthetic Demonstration Generation for Data-Efficient Visuomotor Policy Learning}, 
      author={Zhengrong Xue and Shuying Deng and Zhenyang Chen and Yixuan Wang and Zhecheng Yuan and Huazhe Xu},
      year={2025},
      eprint={2502.16932},
      archivePrefix={arXiv},
      primaryClass={cs.RO},
      note = {\url{https://arxiv.org/abs/2502.16932}}, 
}

@misc{tan2024maniboxenhancingspatialgrasping,
      title={ManiBox: Enhancing Spatial Grasping Generalization via Scalable Simulation Data Generation}, 
      author={Hengkai Tan and Xuezhou Xu and Chengyang Ying and Xinyi Mao and Songming Liu and Xingxing Zhang and Hang Su and Jun Zhu},
      year={2024},
      eprint={2411.01850},
      archivePrefix={arXiv},
      primaryClass={cs.LG},
      note = {\url{https://arxiv.org/abs/2411.01850}}, 
}

@misc{huang2025adversarialdatacollectionhumancollaborative,
      title={Adversarial Data Collection: Human-Collaborative Perturbations for Efficient and Robust Robotic Imitation Learning}, 
      author={Siyuan Huang and Yue Liao and Siyuan Feng and Shu Jiang and Si Liu and Hongsheng Li and Maoqing Yao and Guanghui Ren},
      year={2025},
      eprint={2503.11646},
      archivePrefix={arXiv},
      primaryClass={cs.RO},
      note = {\url{https://arxiv.org/abs/2503.11646}}, 
}

@misc{li2024generalistrobotpoliciesmatters,
      title={Towards Generalist Robot Policies: What Matters in Building Vision-Language-Action Models}, 
      author={Xinghang Li and Peiyan Li and Minghuan Liu and Dong Wang and Jirong Liu and Bingyi Kang and Xiao Ma and Tao Kong and Hanbo Zhang and Huaping Liu},
      year={2024},
      eprint={2412.14058},
      archivePrefix={arXiv},
      primaryClass={cs.RO},
      note = {\url{https://arxiv.org/abs/2412.14058}}, 
}

@misc{black2024pi0visionlanguageactionflowmodel,
      title={$\pi_0$: A Vision-Language-Action Flow Model for General Robot Control}, 
      author={Kevin Black and Noah Brown and Danny Driess and Adnan Esmail and Michael Equi and Chelsea Finn and Niccolo Fusai and Lachy Groom and Karol Hausman and Brian Ichter and Szymon Jakubczak and Tim Jones and Liyiming Ke and Sergey Levine and Adrian Li-Bell and Mohith Mothukuri and Suraj Nair and Karl Pertsch and Lucy Xiaoyang Shi and James Tanner and Quan Vuong and Anna Walling and Haohuan Wang and Ury Zhilinsky},
      year={2024},
      eprint={2410.24164},
      archivePrefix={arXiv},
      primaryClass={cs.LG},
      note = {\url{https://arxiv.org/abs/2410.24164}}, 
}

@misc{ze20243ddiffusionpolicygeneralizable,
      title={3D Diffusion Policy: Generalizable Visuomotor Policy Learning via Simple 3D Representations}, 
      author={Yanjie Ze and Gu Zhang and Kangning Zhang and Chenyuan Hu and Muhan Wang and Huazhe Xu},
      year={2024},
      eprint={2403.03954},
      archivePrefix={arXiv},
      primaryClass={cs.RO},
      note = {\url{https://arxiv.org/abs/2403.03954}}, 
}

@misc{lin2025datascalinglawsimitation,
      title={Data Scaling Laws in Imitation Learning for Robotic Manipulation}, 
      author={Fanqi Lin and Yingdong Hu and Pingyue Sheng and Chuan Wen and Jiacheng You and Yang Gao},
      year={2025},
      eprint={2410.18647},
      archivePrefix={arXiv},
      primaryClass={cs.RO},
      note = {\url{https://arxiv.org/abs/2410.18647}}, 
}

@article{teoh2024green,
  title={Green screen augmentation enables scene generalisation in robotic manipulation},
  author={Teoh, Eugene and Patidar, Sumit and Ma, Xiao and James, Stephen},
  journal={arXiv preprint arXiv:2407.07868},
  year={2024}
}

@misc{mandlekar2023mimicgendatagenerationscalable,
      title={MimicGen: A Data Generation System for Scalable Robot Learning using Human Demonstrations}, 
      author={Ajay Mandlekar and Soroush Nasiriany and Bowen Wen and Iretiayo Akinola and Yashraj Narang and Linxi Fan and Yuke Zhu and Dieter Fox},
      year={2023},
      eprint={2310.17596},
      archivePrefix={arXiv},
      primaryClass={cs.RO},
      note = {\url{https://arxiv.org/abs/2310.17596}}, 
}

@article{yue2024deer,
  title={Deer-vla: Dynamic inference of multimodal large language models for efficient robot execution},
  author={Yue, Yang and Wang, Yulin and Kang, Bingyi and Han, Yizeng and Wang, Shenzhi and Song, Shiji and Feng, Jiashi and Huang, Gao},
  journal={NeurIPS},
  volume={37},
  pages={56619--56643},
  year={2024}
}

@INPROCEEDINGS{10611331,
  author={Xie, Annie and Lee, Lisa and Xiao, Ted and Finn, Chelsea},
  booktitle={ICRA}, 
  title={Decomposing the Generalization Gap in Imitation Learning for Visual Robotic Manipulation}, 
  year={2024},
  volume={},
  number={},
  pages={3153-3160},
  doi={10.1109/ICRA57147.2024.10611331}}

@article{ke20243d,
  title={3d diffuser actor: Policy diffusion with 3d scene representations},
  author={Ke, Tsung-Wei and Gkanatsios, Nikolaos and Fragkiadaki, Katerina},
  journal={arXiv preprint arXiv:2402.10885},
  year={2024}
}

@article{wen2025dexvla,
  title={Dexvla: Vision-language model with plug-in diffusion expert for general robot control},
  author={Wen, Junjie and Zhu, Yichen and Li, Jinming and Tang, Zhibin and Shen, Chaomin and Feng, Feifei},
  journal={arXiv preprint arXiv:2502.05855},
  year={2025}
}

@inproceedings{ma2024hierarchical,
  title={Hierarchical diffusion policy for kinematics-aware multi-task robotic manipulation},
  author={Ma, Xiao and Patidar, Sumit and Haughton, Iain and James, Stephen},
  booktitle={Proceedings of the IEEE/CVF Conference on Computer Vision and Pattern Recognition},
  pages={18081--18090},
  year={2024}
}

@misc{khazatsky2025droidlargescaleinthewildrobot,
      title={DROID: A Large-Scale In-The-Wild Robot Manipulation Dataset}, 
      author={Alexander Khazatsky and Karl Pertsch and Suraj Nair and Ashwin Balakrishna and Sudeep Dasari and Siddharth Karamcheti and Soroush Nasiriany and Mohan Kumar Srirama and Lawrence Yunliang Chen and Kirsty Ellis and Peter David Fagan and Joey Hejna and Masha Itkina and Marion Lepert and Yecheng Jason Ma and Patrick Tree Miller and Jimmy Wu and Suneel Belkhale and Shivin Dass and Huy Ha and Arhan Jain and Abraham Lee and Youngwoon Lee and Marius Memmel and Sungjae Park and Ilija Radosavovic and Kaiyuan Wang and Albert Zhan and Kevin Black and Cheng Chi and Kyle Beltran Hatch and Shan Lin and Jingpei Lu and Jean Mercat and Abdul Rehman and Pannag R Sanketi and Archit Sharma and Cody Simpson and Quan Vuong and Homer Rich Walke and Blake Wulfe and Ted Xiao and Jonathan Heewon Yang and Arefeh Yavary and Tony Z. Zhao and Christopher Agia and Rohan Baijal and Mateo Guaman Castro and Daphne Chen and Qiuyu Chen and Trinity Chung and Jaimyn Drake and Ethan Paul Foster and Jensen Gao and Vitor Guizilini and David Antonio Herrera and Minho Heo and Kyle Hsu and Jiaheng Hu and Muhammad Zubair Irshad and Donovon Jackson and Charlotte Le and Yunshuang Li and Kevin Lin and Roy Lin and Zehan Ma and Abhiram Maddukuri and Suvir Mirchandani and Daniel Morton and Tony Nguyen and Abigail O'Neill and Rosario Scalise and Derick Seale and Victor Son and Stephen Tian and Emi Tran and Andrew E. Wang and Yilin Wu and Annie Xie and Jingyun Yang and Patrick Yin and Yunchu Zhang and Osbert Bastani and Glen Berseth and Jeannette Bohg and Ken Goldberg and Abhinav Gupta and Abhishek Gupta and Dinesh Jayaraman and Joseph J Lim and Jitendra Malik and Roberto Martín-Martín and Subramanian Ramamoorthy and Dorsa Sadigh and Shuran Song and Jiajun Wu and Michael C. Yip and Yuke Zhu and Thomas Kollar and Sergey Levine and Chelsea Finn},
      year={2025},
      eprint={2403.12945},
      archivePrefix={arXiv},
      primaryClass={cs.RO},
      note = {\url{https://arxiv.org/abs/2403.12945}}, 
}

@misc{walke2024bridgedatav2datasetrobot,
      title={BridgeData V2: A Dataset for Robot Learning at Scale}, 
      author={Homer Walke and Kevin Black and Abraham Lee and Moo Jin Kim and Max Du and Chongyi Zheng and Tony Zhao and Philippe Hansen-Estruch and Quan Vuong and Andre He and Vivek Myers and Kuan Fang and Chelsea Finn and Sergey Levine},
      year={2024},
      eprint={2308.12952},
      archivePrefix={arXiv},
      primaryClass={cs.RO},
      note = {\url{https://arxiv.org/abs/2308.12952}}, 
}

@misc{zhong2025surveyvisionlanguageactionmodelsaction,
      title={A Survey on Vision-Language-Action Models: An Action Tokenization Perspective}, 
      author={Yifan Zhong and Fengshuo Bai and Shaofei Cai and Xuchuan Huang and Zhang Chen and Xiaowei Zhang and Yuanfei Wang and Shaoyang Guo and Tianrui Guan and Ka Nam Lui and Zhiquan Qi and Yitao Liang and Yuanpei Chen and Yaodong Yang},
      year={2025},
      eprint={2507.01925},
      archivePrefix={arXiv},
      primaryClass={cs.RO},
      note = {\url{https://arxiv.org/abs/2507.01925}}, 
}

@inproceedings{xing2021kitchenshift,
  title={Kitchenshift: Evaluating zero-shot generalization of imitation-based policy learning under domain shifts},
  author={Xing, Eliot and Gupta, Abhinav and Powers, Sam and Dean, Victoria},
  booktitle={NeurIPS 2021 Workshop on Distribution Shifts: Connecting Methods and Applications},
  year={2021}
}

@misc{tsagkas2025pretrainedvisualrepresentationsfall,
      title={When Pre-trained Visual Representations Fall Short: Limitations in Visuo-Motor Robot Learning}, 
      author={Nikolaos Tsagkas and Andreas Sochopoulos and Duolikun Danier and Sethu Vijayakumar and Chris Xiaoxuan Lu and Oisin Mac Aodha},
      year={2025},
      eprint={2502.03270},
      archivePrefix={arXiv},
      primaryClass={cs.RO},
      note = {\url{https://arxiv.org/abs/2502.03270}}, 
}

@article{zhu2024spa,
  title={Spa: 3d spatial-awareness enables effective embodied representation},
  author={Zhu, Haoyi and Yang, Honghui and Wang, Yating and Yang, Jiange and Wang, Limin and He, Tong},
  journal={arXiv preprint arXiv:2410.08208},
  year={2024}
}

@misc{qu2025spatialvlaexploringspatialrepresentations,
      title={SpatialVLA: Exploring Spatial Representations for Visual-Language-Action Model}, 
      author={Delin Qu and Haoming Song and Qizhi Chen and Yuanqi Yao and Xinyi Ye and Yan Ding and Zhigang Wang and JiaYuan Gu and Bin Zhao and Dong Wang and Xuelong Li},
      year={2025},
      eprint={2501.15830},
      archivePrefix={arXiv},
      primaryClass={cs.RO},
      note = {\url{https://arxiv.org/abs/2501.15830}}, 
}

@article{bu2025agibot,
  title={Agibot world colosseo: A large-scale manipulation platform for scalable and intelligent embodied systems},
  author={Bu, Qingwen and Cai, Jisong and Chen, Li and Cui, Xiuqi and Ding, Yan and Feng, Siyuan and Gao, Shenyuan and He, Xindong and Hu, Xuan and Huang, Xu and others},
  journal={arXiv preprint arXiv:2503.06669},
  year={2025}
}

@misc{yu2021metaworldbenchmarkevaluationmultitask,
      title={Meta-World: A Benchmark and Evaluation for Multi-Task and Meta Reinforcement Learning}, 
      author={Tianhe Yu and Deirdre Quillen and Zhanpeng He and Ryan Julian and Avnish Narayan and Hayden Shively and Adithya Bellathur and Karol Hausman and Chelsea Finn and Sergey Levine},
      year={2021},
      eprint={1910.10897},
      archivePrefix={arXiv},
      primaryClass={cs.LG},
      note = {\url{https://arxiv.org/abs/1910.10897}}, 
}

@misc{zhu20243dgaussiansplattingrobotics,
      title={3D Gaussian Splatting in Robotics: A Survey}, 
      author={Siting Zhu and Guangming Wang and Xin Kong and Dezhi Kong and Hesheng Wang},
      year={2024},
      eprint={2410.12262},
      archivePrefix={arXiv},
      primaryClass={cs.RO},
      note = {\url{https://arxiv.org/abs/2410.12262}}, 
}

@misc{mu2025robotwindualarmrobotbenchmark,
      title={RoboTwin: Dual-Arm Robot Benchmark with Generative Digital Twins (early version)}, 
      author={Yao Mu and Tianxing Chen and Shijia Peng and Zanxin Chen and Zeyu Gao and Yude Zou and Lunkai Lin and Zhiqiang Xie and Ping Luo},
      year={2025},
      eprint={2409.02920},
      archivePrefix={arXiv},
      primaryClass={cs.RO},
      note = {\url{https://arxiv.org/abs/2409.02920}}, 
}

@misc{mu2021maniskillgeneralizablemanipulationskill,
      title={ManiSkill: Generalizable Manipulation Skill Benchmark with Large-Scale Demonstrations}, 
      author={Tongzhou Mu and Zhan Ling and Fanbo Xiang and Derek Yang and Xuanlin Li and Stone Tao and Zhiao Huang and Zhiwei Jia and Hao Su},
      year={2021},
      eprint={2107.14483},
      archivePrefix={arXiv},
      primaryClass={cs.LG},
      note = {\url{https://arxiv.org/abs/2107.14483}}, 
}

@misc{do2025watchlessfeelmore,
      title={Watch Less, Feel More: Sim-to-Real RL for Generalizable Articulated Object Manipulation via Motion Adaptation and Impedance Control}, 
      author={Tan-Dzung Do and Nandiraju Gireesh and Jilong Wang and He Wang},
      year={2025},
      eprint={2502.14457},
      archivePrefix={arXiv},
      primaryClass={cs.RO},
      note = {\url{https://arxiv.org/abs/2502.14457}}, 
}

\clearpage
\appendix

\subsection{Training Details}
\vspace{-3mm}

\label{appendix}

\begin{table}[h]
\caption{Training Details of diffusion policy. We used different training steps in simulation and real environments.}
\centering
\begin{tabular}{ccc}
\toprule
Hyper-parameters          & Simulation & Real-world\\
\midrule
Action Prediction Horizon & 4  & 8\\
Action Horizon & 3 & 8\\
Observation Horizon & 2 & 2\\
Gradient Step / epoch & 220000 & 200\\
batch size                & 128 & 128               \\
Train Denoise Step & 100 & 50\\
Val Denoise Step & 10 & 10 \\

\bottomrule
\end{tabular}
\vspace{-3mm}
\end{table}

\vspace{-2mm}

\begin{table}[h]
\caption{
The results presented in~\cref{table:simulation} primarily use a dataset of 20k timesteps. For particularly simple or challenging tasks, smaller (5k) or larger (100k) datasets are used accordingly.
}
\label{tab:train_data}
\centering
\begin{tabular}{l|l@{\hspace{3mm}}l@{\hspace{1mm}}lll}
\toprule
Task & Pick & Push & Box & Assembly & Hammer\\
\midrule
\# Timesteps & 20k  & 20k & 100k & 20k & 20k\\
\toprule
Task & Wall & Window & Faucet & Disassembly & Basketball\\
\midrule
\# Timesteps & 20k  & 5k & 5k & 20k & 20k\\

\bottomrule
\end{tabular}
\end{table}

\vspace{-2mm}

\subsection{Data collection and Scoring Details}
\vspace{-1mm}

\label{app:real}

In this subsection, we provide a detailed introduction to data collection and the scoring criteria during evaluation.

\begin{figure}[!htbp] 
\centering
\includegraphics[width=0.45\textwidth]{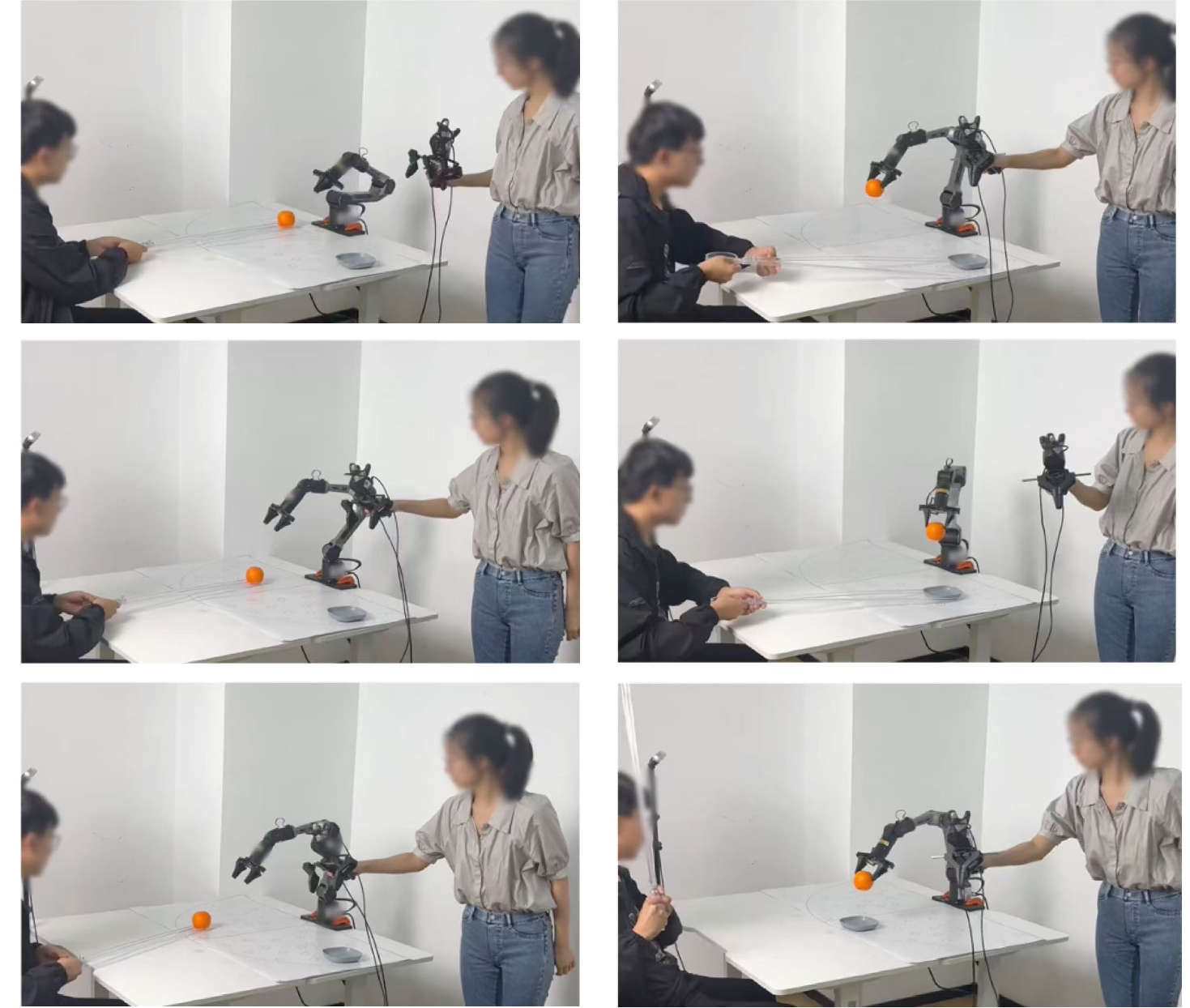}
\caption{
An example of the \emph{MOVE} data collection paradigm. Captured using an external smartphone for better visibility, rather than the camera used for data collection.
}
\label{fig:collect}
\vspace{-2mm}
\end{figure}

\paragraph{Data Collection}  
In the static data collection setup, Operator 1 performs teleoperation by Pika teleoperation device while Operator 2 positions the objects manually. In contrast, the \emph{MOVE} paradigm retains the same teleoperation procedure but extends Operator 2's role to include dynamic object manipulation. 
Specifically, Operator 2 employs a 3D-printed transparent resin tongs to move oranges or plates during each trajectory. 
Note that Operator 2 remains outside the camera’s field of view to avoid interfering with the visual input used for policy training.
A visual overview of the real-world data collection process is shown in Figure~\ref{fig:collect}.

The simulation data collection script has a small chance of failure, and we rerun it as needed until the desired amount of data is collected.

\begin{table}[t]
\centering
\caption{
Comparison of static and \emph{MOVE} collection paradigm.
}
\label{table:length}
\begin{tabular}{ccccc}
\toprule
\multirowcell{3}{Metrics}& \multicolumn{2}{c}{Real-world} & \multicolumn{2}{c}{Simulation}\\
\cmidrule(lr){2-3} \cmidrule(lr){4-5}
  & \multirowcell{2}{Average \\timesteps} & \multirowcell{2}{Personnel\\Required}  & \multirowcell{2}{Average \\timesteps} & \multirowcell{2}{Generation \\successful rate} \\
& & & &\\
\midrule
Static & 459.2 & 2 & 83.8 & 93.3\% \\
\emph{MOVE} & 549.3 & 2 & 107.4 & 89.7\% \\
\bottomrule
\end{tabular}
\vspace{-4mm}
\end{table}

\vspace{-2mm}

\begin{figure}[h] 
\centering
\includegraphics[width=0.35\textwidth]{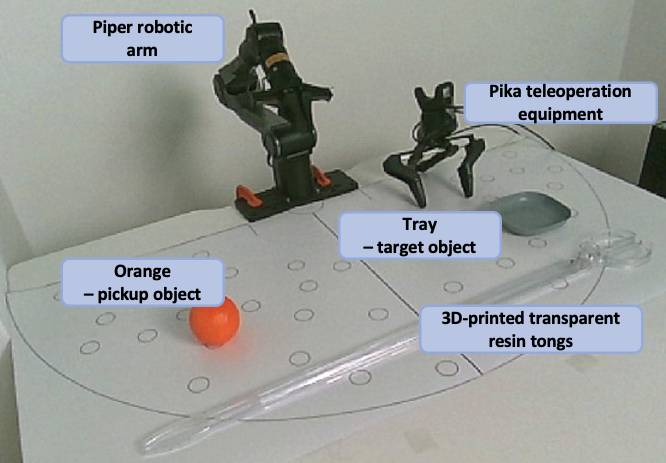}
\caption{
An overview of our real-world experimental setup, captured by a RealSense camera.  
We sample pairs of random points on the table as placement locations for the oranges and trays, which serve as training data.  
Object movements are performed using 3D-printed transparent resin tongs.
}
\label{fig:desk}
\vspace{-2mm}
\end{figure}

\paragraph{Scoring Criteria} Following~\cite{lin2025datascalinglawsimitation}, we report the normalized score as a metric for the robot manipulation to concretely evaluate the capability of the policy step by step.
\begin{itemize}
    \item 0 points: The gripper does not move toward the orange or moves around it without any contact.
    \item 1 point: The gripper touches the orange but does not grasp it due to minor errors.
    \item 2 points: The gripper successfully grasp the orange.
    \item 3 points: The gripper put the orange on the tray.
\end{itemize}

\end{document}